\def\isarxiv{1}

\newif\ifarxiv
\arxivfalse
\ifdefined\isarxiv
  \arxivtrue
\fi

\ifarxiv
  \documentclass[sigconf, nonacm]{acmart}
\else
\documentclass[sigconf, anonymous, nonacm, review]{acmart}
 \fi

\usepackage{multirow} \usepackage{algorithm}

\usepackage{placeins} 

\usepackage[noend]{algpseudocode}

\usepackage{booktabs}
\usepackage{graphicx}  \usepackage[]{mdframed}

\newcommand{\stitle}[1]{\vspace{1ex}\noindent\textbf{#1}}
\usepackage{enumitem}
\setlist{noitemsep,topsep=0pt,parsep=0pt,partopsep=0pt, leftmargin=*}

\usepackage{tcolorbox}
\usepackage{xcolor, soul}
\sethlcolor{lightgray}

\usepackage{pifont}   \newcommand{\xmark}{\ding{55}}

\usepackage{tabularx}
\usepackage{subcaption}

\usepackage{xspace}
\newcommand{\sysname}{\textsc{Bonsai}\xspace}

\newcounter{VasiaNOC}
\stepcounter{VasiaNOC}

\newcounter{YuhangNOC}
\stepcounter{YuhangNOC}

\newcounter{RomaricNOC}
\stepcounter{RomaricNOC}

\newcounter{NaimaNOC}
\stepcounter{NaimaNOC}

\begin{document}
\ifarxiv
  \title{Not All Neighbors Matter: Understanding the Impact of Graph Sparsification on GNN Pipelines}
\else
  \title{Not All Neighbors Matter: Understanding the Impact of Graph Sparsification on GNN Pipelines: [Experiments \& Analysis]}
\fi

\author{Yuhang Song}
\affiliation{\institution{Boston University}
  \city{Boston}
  \state{MA}
  \country{USA}
}
\email{yuhangs@bu.edu}

\author{Naima Abrar Shami}
\affiliation{\institution{Boston University}
  \city{Boston}
  \state{MA}
  \country{USA}
}
\email{anaima@bu.edu}

\author{Romaric Duvignau}
\affiliation{\institution{Chalmers University of Technology}
  \city{Gothenburg}
  \country{Sweden}
}
\email{duvignau@chalmers.se}

\author{Vasiliki Kalavri}
\affiliation{\institution{Boston University}
  \city{Boston}
  \state{MA}
  \country{USA}
}
\email{vkalavri@bu.edu}

\begin{abstract}
As graphs scale to billions of nodes and edges, graph Machine Learning workloads are constrained by the cost of multi-hop traversals over exponentially growing neighborhoods. While various system-level and algorithmic optimizations have been proposed to accelerate Graph Neural Network (GNN) pipelines, data management and movement remain the primary bottlenecks at scale. In this paper, we explore whether \emph{graph sparsification}, a well-established technique that reduces edges to create sparser neighborhoods, can serve as a lightweight pre-processing step to address these bottlenecks while preserving model quality.

We develop \sysname, an extensible experimental framework that integrates graph sparsification into GNN training and inference pipelines, and use it to study how four representative sparsification methods interact with four GNN architectures, across six graphs and two learning tasks (node classification and link prediction). Our central finding is counter-intuitive: \emph{removing a large fraction of a graph's edges, often the majority, has only a small effect on model quality, and sometimes improves it}. For example, the K-Neighbor sparsifier removes 83.3\% of the edges of the Products graph while GCN accuracy drops by only 0.2 percentage points, and on the Arxiv link prediction task, sparsification raises GraphSAGE test AUC from 0.86 to 0.95. Because the runtime costs scale with the number of edges, these reductions translate into substantial speedups. Our results show that the pre-processing overhead is quickly amortized, making sparsification practical even for very large graphs.
\end{abstract}

\maketitle

\section{Introduction}

Graph Neural Network (GNN) workloads~\cite{pinsage, pickandchoose, zhang2023survey, Yasmin2025, schlichtkrull2018modeling, nguyen2021graphdta, streaminggnn} become bottlenecked by irregular memory access, high feature I/O, and the \emph{neighborhood explosion} that arises when traversing GNN layers of graphs with billions of nodes and edges.  Various systems and algorithmic approaches have been proposed to address these issues: distributed training~\cite{song2023adgnn, wan2023scalable, aineutrontp, zheng2020distdgl, zheng2024graphstorm}, multi-GPU data- and model-parallel pipelines~\cite{daha, bytegnn, liu2023bgl}, out-of-core storage~\cite{ginex, diskgnn, surveyoocgnn, smartssdsampler}, specialized data structures~\cite{lotan2023, waleffe2023mariusgnn}, indexing and prefetching to reduce random-access latency~\cite{reordering, ringsampler}, and algorithmic modifications~\cite{rong2020dropedge, huang2021scaling}. Yet comprehensive evaluations~\cite{yuan1comprehensive} show that even with these optimizations,  \emph{data management and movement remain the primary bottleneck for large-scale GNNs}.  These overheads are ultimately driven by the number of edges, which directly affects the cost of loading, sampling, and aggregating neighborhoods.

Motivated by recent evidence that random sampling can improve generalization and reduce training costs~\cite{vatter2024size,bajaj2024graphneuralnetworktraining}, in this paper, we ask a fundamental question:
\emph{how much of the graph structure is actually necessary for effective learning?} Our intuition is that real-world graphs are noisy, redundant, and often exhibit heavy-tailed degree distributions. As a result, many edges may be structurally redundant for the downstream learning objective. To this end, we explore if \emph{graph sparsification}~\cite{localdegree,rankdegree,kneighbor,chen2023demystifying}, a standard data management technique, can be applied as a lightweight pre-processing step to accelerate GNN training and inference workloads. Instead of scaling the system or modifying the learning algorithm, we consider compressing the graph’s structure before learning, with the goal of reducing memory, I/O, and neighborhood sampling overhead, while preserving predictive accuracy.  As a result,  a one-off sparsification step could benefit every subsequent training and inference run, with no change to the model or the pipeline.

Despite its intuitive appeal, the impact of graph sparsification on modern GNNs remains poorly understood. Prior work~\cite{vatter2024size, kosman2022lsp, luo2021learning, shen2024graph, zheng2020robust} typically evaluates individual sparsification techniques in isolation, focuses on a single model, or restricts experiments to small datasets. To study this problem under consistent conditions, we develop an extensible framework that integrates graph sparsification into GNN  pipelines, and we use it to evaluate how four representative sparsification methods interact with common GNN architectures across graph scales and two learning tasks.  Our aim is not to identify a single best method, but to provide an extensible tool that helps practitioners and researchers understand when sparsification is worthwhile.
 \vspace{1mm}

\noindent We make the following contributions:
\begin{itemize}

   \item We develop \sysname, an extensible experimental framework that is compatible with DGL~\cite{wang2019dgl} and allows users to transparently integrate graph sparsification as a pre-processing step in their GNN training and inference pipelines (\S\ref{sec:exp-framework}). Currently, \sysname supports four graph sparsification techniques (\emph{Random}, \emph{K-Neighbor}, \emph{Rank Degree}, \emph{Local Degree}), four widely-used GNN architectures (\emph{GCN}, \emph{GraphSAGE}, \emph{GAT}, \emph{SGFormer}), and six real-world graphs, of different scales, domains, and homophily levels (\emph{PubMed}, \emph{CoauthorCS}, \emph{Roman-empire}, \emph{Arxiv}, \emph{Products}, \emph{Papers100M}).
        
    \item We define a structured suite of evaluation metrics that quantify how graph sparsification affects accuracy–efficiency trade-offs across models and datasets. Our evaluation goals cover training dynamics, serving-time behavior, pre-processing overhead, and the sensitivity of compression levels and accuracy to various graph sparsification parameters (\S\ref{sec:methodology}). We evaluate these metrics on both node classification
and link prediction.
        
    \item We present and analyze the first set of results collected from \sysname, revealing how graph sparsification impacts GNN accuracy, training convergence, and end-to-end performance, providing practical guidance on when compression is a viable alternative or complement to systems-level scaling (\S\ref{sec:results}).
    
\end{itemize}

\begin{figure*}[t]
    \centering
    \includegraphics[width=\textwidth]{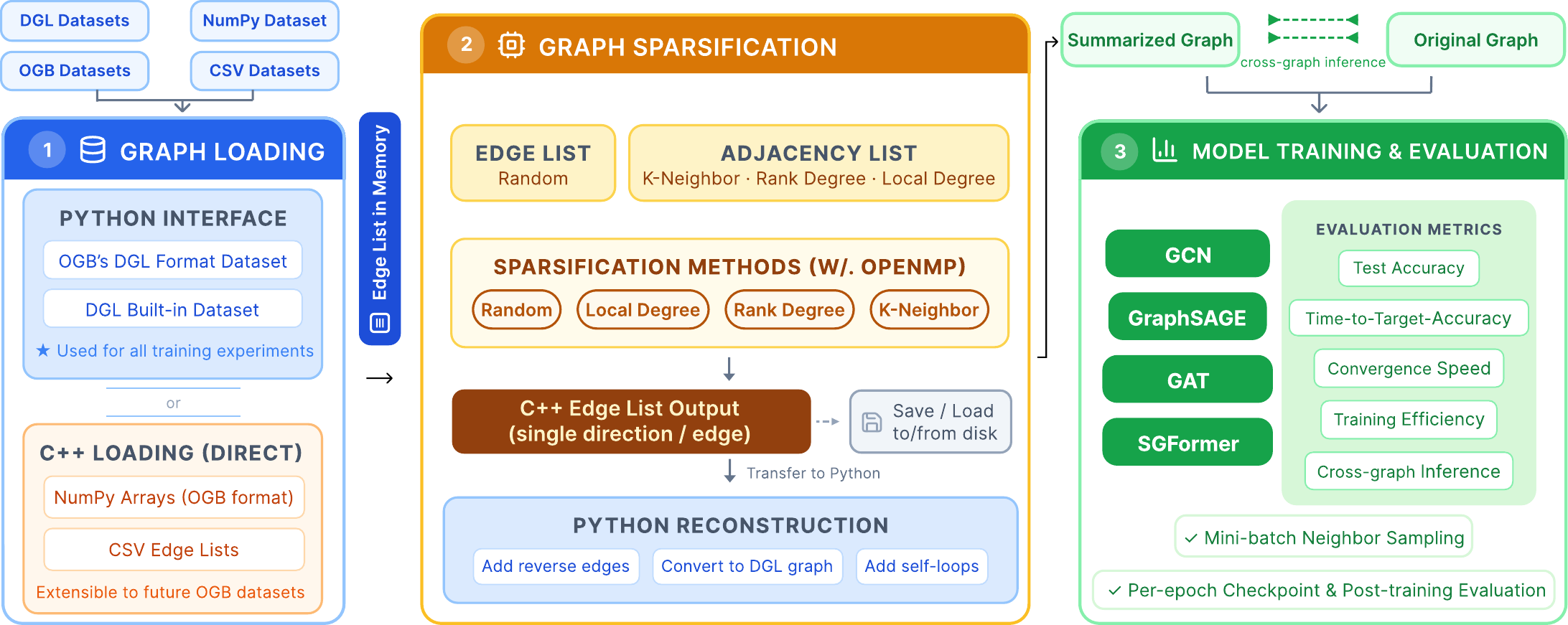}\vspace{-2mm}
    \caption{Overview of \sysname}
    \label{fig:framework-overview}
\end{figure*}

\stitle{Summary of major findings.} Our central result is that \textbf{graph reduction can often preserve, and sometimes even improve, predictive performance}. Across all evaluated GNN architectures, there exists at least one instance of a model trained on a sparsified graph that exceeds the accuracy of the same model trained on the original graph. 
Strikingly, this holds even for the simplest strategy: dropping edges uniformly at random degrades accuracy gracefully and produces some of the largest gains we observe, raising GAT's accuracy on PubMed from 0.73 to 0.78 and its link-prediction AUC on Arxiv from 0.63 to 0.81. 
Degree-based methods push the reduction much further.  \textbf{The K-Neighbor sparsifier consistently achieves a particularly strong trade-off between efficiency and accuracy}, across datasets and models. When evaluated with the GraphSAGE model on Products, it reaches the accuracy of a model trained on the original graph (within 1\%) up to $8\times$ faster.  The clear exception is Rank Degree, the most aggressive structure-aware method, which is  the \emph{worst} of the four, as it discards so much of the graph that accuracy collapses on the larger datasets.  With regard to performance, we find that \textbf{the benefits of reduction become more pronounced at scale}, substantially accelerating both training and inference.  Importantly, we also observe that \textbf{the computational overhead of sparsification is quickly amortized}, making it a practical pre-processing step even for very large graphs. Finally, \textbf{these findings are not limited to node classification}: the same patterns carry over to link prediction, including on heterophilous graphs. \vspace{1mm}

\noindent \sysname and all data and experimental results of this paper will be made open-source upon paper acceptance.

 \section{Experimental Framework}\label{sec:exp-framework}

We build \sysname, a generic experimental framework that seamlessly integrates high-performance C++ implementations of graph sparsification methods with Python-based DGL~\cite{wang2019dgl} pipelines. The overall training pipeline is illustrated in Figure~\ref{fig:framework-overview} and consists of three main components: (1) graph loading, (2) graph sparsification, and (3) model training and evaluation.

\subsection{Framework design and implementation}

\stitle{Graph loading.} Users can load their graphs into \sysname using either our direct C++ interface or a Python interface. The C++ interface reads the NumPy/CSV formats used by OGB~\cite{hu2021opengraphbenchmarkdatasets}, enabling direct loading of large datasets such as Arxiv, Products, and Papers100M, while the Python interface supports OGB and DGL datasets. To ensure fair comparisons, we use the Python data loading interface for all experiments in this paper.

\stitle{Graph sparsification.} Once the graph data is loaded, it is represented in an edge-list format. Existing GNN literature \cite{kipf2017gcn, shchur2019pitfallsgnn, hamilton2017graphsage, huang2021scaling}, as well as DGL's built-in dataset loaders, treat graphs as undirected by adding reverse edges and self-loops. We follow the same convention: we store edge-list representation as provided by the dataset loader and add reverse edges when converting to adjacency lists for structure-aware sparsification methods.

Different sparsification methods require different input representations. For example, random sparsification operates directly on the edge list, while K-Neighbor performs graph traversal and operates more naturally on adjacency-list representations. \sysname provides C++ conversion methods to handle such cases. Since this conversion step is method-dependent, we include its cost as part of the sparsification time and report it in the results (\S\ref{sec:q4}).

\stitle{Model training and evaluation.}  Model training is performed using either mini-batch neighbor sampling or full-graph training, depending on the model. We implement four representative GNN architectures: GCN, GAT,  GraphSAGE, and the SGFormer graph transformer. By keeping the training setup consistent across models, we ensure that observed performance differences arise from graph structure changes rather than model-level tuning. While node classification is our primary task, \sysname also supports \emph{link prediction}: the same GNN encoders produce node embeddings, positive edges are scored against randomly sampled negative edges, and model quality is measured by test AUC (cf.~\S\ref{sec:q1}).

We decouple training from testing by checkpointing model weights after each epoch and performing post-hoc evaluation. This enables accurate reconstruction of accuracy trajectories over time and allows us to compute time-to-accuracy metrics without introducing evaluation overhead during training. In addition to final accuracy (or AUC for link prediction), we report convergence time and training efficiency, which are central to understanding the practical benefits of graph sparsification (cf.~\S\ref{sec:q2}).

To further analyze the robustness of learned representations, \sysname supports \emph{cross-graph inference}: models trained on the original graph can be evaluated on summarized graphs, and vice versa. This allows us to assess whether sparsification preserves not only accuracy but also transferable structural information relevant for downstream inference.

\stitle{Support for large-scale graphs.} For datasets that exceed single-GPU or single-machine memory limits, we provide a separate training pipeline based on DGL’s GraphBolt backend \cite{wang2019dgl}. In this setting, original or sparsified graphs are loaded from disk in a streaming fashion, and node features are accessed via memory-mapped storage. This pipeline enables end-to-end experiments on billion-edge graphs, such as Papers100M.

\stitle{Extensibility and reproducibility.} \sysname is designed to be easily extensible, supporting the seamless addition of new sparsification methods, datasets and GNN models.  A new sparsifier is added by registering an entry in the \texttt{SummarizationMethod} \texttt{enum} and implementing it behind the unified \texttt{summarize\_graph} entry point, with no changes to the Python training pipeline.  Integrating new GNN models is equally straightforward through a standard forward interface that consumes a DGL message-flow graph and node features.  Since sparsification operates solely on the graph structure prior to training, no model-level changes are needed to train on summarized graphs.

For reproducibility, all experiments use global seeding across Python, NumPy, PyTorch, and DGL,  and all stochastic sparsifiers accept explicit seeds. Dataset and model-specific hyperparameters can be specified via configuration files, ensuring consistent settings across large experiment sweeps.  Finally,  all experimental metadata and checkpoints are logged with Weights \& Biases~\cite{wandb},  which records detailed environment information and enables exact reproduction and post-hoc result analysis.

 \subsection{Sparsification methods}\label{sec:summarization}

For the purpose of this work, we select four representative sparsification techniques from a recent study~\cite{chen2023demystifying} that cover a range of strategies for reducing graph size while attempting to preserve structural properties: \textbf{Random Sparsifier}, \textbf{K-Neighbor Sparsifier}, \textbf{Rank Degree Sparsifier}, and \textbf{Local Degree Sparsifier}. Each method has different assumptions and design choices, which could lead to different trade-offs between efficiency and accuracy. Importantly, all four methods are lightweight and amenable to parallelization, meaning their computational cost can be easily amortized in a large-scale training pipeline.  We deliberately exclude other reduction techniques (e.g.,  Hashemi et al.~\cite{hashemi2024comprehensivesurveygraphreduction} provide a comprehensive survey) that are either closely related to the simpler degree-based strategies we evaluate (e.g., DSpar~\cite{liu2023dspar}) or less lightweight and not purely structural (e.g., TADA~\cite{lai2024efficientta}). We describe each selected method below and provide pseudocode in Appendix~\ref{app:pseudocode}.

\stitle{Random}. This method reduces the size of a graph by independently retaining each edge with a fixed probability $p$.  We implement this efficiently in a parallel fashion,  converting the input graph to an edge list, then uniformly sampling at random to achieve the desired sparsification ratio. We run the sampling directly on the original dataset, if the dataset has both directions of an edge, each direction is sampled independently, so an edge is removed from the undirected graph only if \emph{both} of its directed copies are removed.

\stitle{K-Neighbor}. This sparsifier constructs a reduced graph by selecting up to $k$ incident edges for each vertex~\cite{kneighbor}. For a vertex $i$, let $N(i)$ denote its set of neighbors and $d_i = |N(i)|$ denote its degree. If $d_i \le k$, all incident edges are retained. Otherwise, $k$ neighbors are sampled uniformly at random without replacement from $N(i)$. For undirected graphs, if an edge is selected by either endpoint, we consider it as selected. This method guarantees that each vertex preserves at least $\min(k, d_i)$ incident edges, making it suitable for downstream tasks that require sufficient local connectivity. However, this design may also impact the extent of edge removal, since an edge is retained if either of its endpoints selects it.

\stitle{Rank Degree}. This method starts with selecting a random set of “seed” vertices, then iteratively adds top $\rho$ neighboring vertices based on their degree rank until a target vertex size $x$ is reached~\cite{rankdegree}. In our experiments, we use the training nodes as the initial seed nodes. Rank Degree aims to preserve important structural properties of the graph by focusing on high-degree nodes, which are often more influential in GNN training.
This sparsification method is inherently sequential, as each iteration depends on the previous one. We implement it in an iterative fashion, where in each hop we select neighbors of the current vertex set by degree rank, update the selected set, and continue until the target size is reached. This parallelizes selection within each hop while preserving the algorithm's sequential structure. We make one modification to avoid removing selected edges between iterations, which is prohibitively expensive at scale. Instead, we leave them in place and deduplicate afterward, so correctness is preserved despite possible resampling. The algorithm also includes a fallback that re-seeds with random nodes when seeds are exhausted, but this never occurred in our experiments, as the high average degree of our graphs ensures each iteration yields enough new seed candidates.

\stitle{Local Degree}. For each node $i \in V$, the Local Degree sparsifier selects the edges to the top $\operatorname{d}(i)^\alpha$  neighbors, sorted by degree in descending order~\cite{localdegree}. Here, $\alpha$ is a parameter that controls the sparsification level. The goal of this approach is to keep those edges in the sparsified graph that lead to nodes with high degree.
Similarly to K-Neighbor Sparsifier, when running on an undirected graph, an edge is retained if either of its endpoints selects it.

\begin{table*}[t]
\centering
\small
\caption{Characteristics of the sparsifiers we consider in this study.  All methods are edge sparsifiers that preserve the full vertex set $V$ but they differ in what controls the amount of reduction and how they reshape the graph's structural properties (\S\ref{sec:results}).}
\vspace{-2mm}
\renewcommand{\arraystretch}{1.2}
\setlength{\tabcolsep}{4pt}
\begin{tabularx}{\textwidth}{@{}p{1.6cm}X p{2.4cm} p{2.4cm} c X@{}}
\toprule
\textbf{Method} & \textbf{Edge selection rule} & \textbf{Structure used} & \textbf{Reduction driver} & \textbf{Per-node connectivity} & \textbf{Effect on degree distribution} \\
\midrule
\textbf{Random} &
Each edge kept i.i.d.\ with prob.\ $p$ &
None (structure-agnostic) &
Edge retention prob.\ $p$ &
\xmark{} (probabilistic) &
Shape preserved, uniformly scaled by $p$ \\
\addlinespace
\textbf{K-Neighbor} &
Up to $k$ incident edges per node (uniform among neighbors) &
Local degree only &
Per-node edge cap $k$ &
\checkmark{} ($\min(d_i,k)$ edges) &
\emph{Caps} degree at $\approx\!k$; flattens the tail \\
\addlinespace
\textbf{Local Degree} &
Edges to the top $d_i^{\alpha}$ \emph{highest-degree} neighbors of each node &
Local + neighbor degree &
Per-node edge budget ($\alpha$) &
\checkmark{} ($\geq 1$ edge if $d_i\!>\!0$) &
Hub-biased; preserves high-degree backbone \\
\addlinespace
\textbf{Rank Degree} &
BFS from seed nodes, expanding to top-$\rho$ highest-degree neighbors, keeping only edges of selected nodes &
Global, seed-dependent, iterative &
Target \#\,of \emph{selected vertices} $x$ &
\xmark{} (non-selected nodes isolated) &
Isolates non-selected nodes; keeps a small core of hubs \\
\bottomrule
\end{tabularx}
\label{tab:sparsifier_comparison}
\end{table*} 
Table~\ref{tab:sparsifier_comparison} summarizes the key characteristics of the four methods that drive their behavior in our experiments.  While all are edge sparsifiers that preserve the full vertex set and only drop edges,  they differ in what controls the amount of reduction. Random, K-Neighbor, and Local Degree control an \emph{edge} budget (a retention probability or a per-node edge cap), so every node keeps a share of its edges. Rank Degree instead grows a set of vertices from its seeds until it reaches a target \emph{vertex count}, and keeps only the edges among those selected vertices; as a consequence, the vertices it never selects are left isolated, with no edges at all. A second, related distinction is whether a method offers a \emph{per-node connectivity guarantee}: K-Neighbor and Local Degree always retain at least one incident edge per (originally non-isolated) node, while Random and especially Rank Degree may leave nodes with no edges. As we show in \S\ref{sec:results}, these two properties largely explain why some methods preserve accuracy while others degrade it. \section{Evaluation Methodology}\label{sec:methodology}
In this section, we first define the evaluation metrics that we use to quantify how graph sparsification affects accuracy–efficiency trade-offs across models and datasets (\S\ref{subsec:eval_goals}). Then, we introduce the datasets (\S\ref{sec:datasets}), models, and hyperparameter configuration (\S\ref{sec:models}) we use across all experiments. Finally, we describe the default parameters we use for each sparsification algorithm (\S\ref{sec:spars-params}).

\subsection{Evaluation goals and metrics} \label{subsec:eval_goals}
Our study aims to answer the following research questions:

\stitle{[Q1] Accuracy and time to convergence}: \emph{How does graph sparsification affect the best accuracy a GNN model can achieve, and the time needed to reach it?}\vspace{1mm} 

\noindent To answer this question, we measure two metrics. (i)~\textbf{Maximum accuracy} is the highest test accuracy a model can reach under our uniform stopping rule, where training stops after 50 consecutive epochs without validation improvement or after 2000 total epochs. We report maximum accuracy separately on the original and sparsified graphs to isolate any loss or gain in a model's representational capacity due to graph compression. (ii)~\textbf{Time to convergence} captures the wall-clock time for a model to reach its own maximum accuracy on a given graph (original or sparsified). Together, these metrics reveal whether sparsification alters the best achievable accuracy and whether it accelerates or slows down training.

\stitle{[Q2] Training efficiency}: \emph{Does sparsification reduce the time to reach a target accuracy?}\vspace{1mm}

\noindent We address this question by measuring \textbf{time-to-target accuracy}, computed by taking a model's maximum test accuracy on the original graph, with a 1\% tolerance margin, as a target and measuring the training time required for the same model to reach at least that test accuracy on the sparsified graph. If the target is not reached before early stopping, we report a timeout. This metric indicates whether sparsification can match the original graph's best performance in less time. We further report per-epoch accuracy over training time to reveal how the convergence behavior of a model is affected by sparsification.

\stitle{[Q3] Serving-time trade-offs}: \emph{Given a model trained on the original graph, can we accelerate serving using the sparsified graph while preserving accuracy?}\vspace{1mm}

\noindent For this question, we examine deployment-time trade-offs when models are trained on the original graph but inference is performed on a summarized one. This is of interest because GNN inference is \emph{stateful}. That is, the graph structure and features used during training must be accessible at inference time. Although model parameters typically fit in GPU memory, scalability is often hindered by loading and transferring node neighborhoods and features across the memory hierarchy. With parameters determined from training on the original graph, we run forward-only inference on the summarized graph and measure \textbf{inference runtime} to quantify potential serving speedups due to operating on a smaller graph and \textbf{inference accuracy} to assess potential fidelity loss due to the compressed representation.

\begin{table*}[t]
\centering
\small
\caption{Graph datasets we use in this study and their structural properties}\vspace{-3mm}
\renewcommand{\arraystretch}{1.1}
\setlength{\tabcolsep}{4pt}
\begin{tabular}{l|cccc|cccc}
\toprule
\textbf{Dataset} & \textbf{Nodes} & \textbf{Edges} & \textbf{Feature Dim.} & \textbf{File Size} & \textbf{Avg.\ deg.} & \textbf{Max deg.} & \textbf{Deg.\ Gini} & \textbf{Edge Homophily} \\
\midrule
PubMed       & 19.7K   & 44.3K   & 500  & 54.5 MB   & 4.5  & 171       & 0.60 & 0.80 \\
CoauthorCS   & 18.3K   & 81.9K   & 6805 & 477.8 MB  & 8.9  & 136       & 0.46 & 0.81 \\
Roman-empire & 22.7K   & 32.9K   & 300  & 19.5 MB   & 2.9  & 14        & 0.18 & 0.05 \\
Arxiv        & 169.3K  & 1.2M    & 128  & 79.2 MB   & 13.7 & 13{,}161  & 0.63 & 0.65 \\
Products     & 2.4M    & 61.9M   & 100  & 1.4 GB    & 50.5 & 17{,}481  & 0.62 & 0.81 \\
Papers100M   & 111.1M  & 1.6B    & 128  & 56.2 GB   & 29.1 & 251{,}471 & 0.67 & 0.57 \\
\bottomrule
\end{tabular}
\label{tab:dataset_combined}
\end{table*}

\stitle{[Q4] Pre-processing overhead}: \emph{What is the upfront cost of sparsification and how does it compare to downstream savings?}\vspace{1mm}

\noindent To evaluate the pre-processing overhead and understand potential end-to-end benefits, we measure the \textbf{sparsification time} as the wall-clock time from reading the original graph to emitting the summarized result, without training included. We also report whether a sparsification algorithm pays for itself in a single training run by comparing the end-to-end training time on the original graph to that on the sparsified graph, including pre-processing costs.

\stitle{[Q5] Compression vs. performance}: \emph{How do reductions in graph size translate into runtime gains?}\vspace{1mm}

\noindent We answer this by reporting the \textbf{edge reduction percentage} a sparsifier achieves relative to test accuracy and training time.  We measure \emph{unique undirected edges} (equal to the relative reduction in message-passing edges), as a proxy for memory and I/O requirements,  to help us explain the observed time savings. We further compare methods at matched compression levels, decoupling edge-selection quality from the amount of graph retained.\vspace{1mm}

While we state the above questions in terms of classification accuracy, we also answer them for the \emph{link prediction} task, measuring model quality with the AUC: each test edge is scored against a randomly sampled non-edge, and the AUC is the probability that the model ranks the true edge above the negative one, so $0.5$ is random guessing and $1.0$ perfect ranking~\cite{kipf2016variationalgraphautoencoders}. For this task, we evaluate every dataset and method except Papers100M, which is prohibitively slow at this scale on our hardware; in its place we add Roman-empire, a \emph{heterophilous} graph in which connected nodes tend to have \emph{different} labels.  We describe all datasets next.

\subsection{Datasets}\label{sec:datasets}
We select six widely-used real-world graphs that include (i) two small-scale graphs with simple features and labels, (ii) two medium-to-large-scale OGB graphs with standardized splits, (iii) one web-scale graph, and (iv) one heterophilous graph in which connected nodes rarely share the same label. Our goal is to provide experimental results on graphs from diverse domains and cover a spectrum of degree distributions, feature dimensionalities, homophily levels, and graph sizes. Specifically, we use the following datasets:

\begin{itemize}
    \item \textbf{PubMed (small)}~\cite{pubmed} is a classic benchmark with sparse TF-IDF features and three classes \cite{zhang2023survey}. This dataset has been widely used in prior studies~\cite{zhang2023survey,jia2020roc,bajaj2024graphneuralnetworktraining}. 
    \item \textbf{CoauthorCS (small–medium)} is a feature-rich co-authorship network with 15 classes and higher average degree than PubMed. It is based on the Microsoft Academic Graph from the KDD Cup 2016 challenge ~\cite{sinha2015overview}.
     \item \textbf{Roman-empire (small-medium, heterophilous)}~\cite{platonov2023critical} is a word-dependency graph in which adjacent nodes rarely share a label.  Unlike the other datasets, it exhibits very low homophily and a chain-like, low-degree structure. We include it to test whether our sparsification findings generalize to the heterophilous setting, where neighboring nodes seldom share labels.
    \item \textbf{Arxiv (medium)}~\cite{hu2021opengraphbenchmarkdatasets} is a graph representing the citation network between all Computer Science arXiv papers indexed by MAG. Each node is an arXiv paper and each edge indicates that one paper cites another.
    \item \textbf{Products (large)}~\cite{hu2021opengraphbenchmarkdatasets} is a product co-purchasing graph with millions of nodes and tens of millions of edges, also widely used in prior studies and GNN works~\cite{bajaj2024graphneuralnetworktraining, wu2023sgformer, ringsampler, liu2023bgl}.
    \item \textbf{Papers100M (very large)}~\cite{hu2021opengraphbenchmarkdatasets} is a citation graph with over 100M nodes and 1.6B edges. We use this graph to evaluate whether sparsification can make training and inference feasible or substantially faster at large scales.
\end{itemize}\vspace{1mm}

\noindent Table~\ref{tab:dataset_combined} summarizes the size characteristics and structural properties of these datasets. For PubMed and CoauthorCS, we report the dataset size as the combined size of the undirected edge list file and the feature file. For the rest, we report the size of the original dataset file downloaded.  Properties like the average degree, the skew of the degree distribution (its Gini coefficient), and the edge homophily determine how much a graph can be sparsified and can help explain how each method preserves model quality.

\subsection{GNN models and configuration}\label{sec:models}

We study the effects of sparsification on four widely-used GNN architectures, and in particular, we focus on the task of \emph{node classification}, complemented by a \emph{link prediction} experiment in \S\ref{sec:q1}: Graph Convolutional Networks (GCN)~\cite{kipf2017gcn}, Graph Attention Networks (GAT)~\cite{velivckovic2017gat}, GraphSAGE~\cite{hamilton2017graphsage}, and graph transformers (SGFormer)~\cite{wu2023sgformer}. 
These models represent a diverse set of approaches to graph representation learning, allowing us to evaluate the effectiveness of sparsification techniques across different architectures. We note that GCN, GAT, and GraphSAGE are canonical message-passing GNNs and form the primary basis for our conclusions, while SGFormer provides only a partial view of transformer-style models due to the resource constraints described below.

We use \textbf{GCN}~\cite{kipf2017gcn} (spectral convolution), \textbf{GAT}~\cite{velivckovic2017gat} (attention-based message passing), and \textbf{GraphSAGE}~\cite{hamilton2017graphsage} (inductive neighbor sampling) as canonical message-passing GNNs, chosen to cover convolution, attention, and sampling. We additionally include \textbf{SGFormer}~\cite{wu2023sgformer}, a graph transformer that combines a GNN branch and a global-attention branch on the full graph. As a transformer model, SGFormer has much higher memory requirements than the other models we consider in this study; we therefore implement it in DGL and train it with mini-batch neighbor sampling, adopting the sampling parameters used for the larger datasets in the original paper~\cite{wu2023sgformer, sgformer-repo} (cf.~Table~\ref{tab:hyperparameters} in Appendix~\ref{app:hyperparameters}), which allows us to evaluate it across all datasets, including Papers100M.

\stitle{Hyperparameter configuration.}
We tune hyperparameters per dataset and model (details in Appendix~\ref{app:hyperparameters}). For all models on PubMed, Arxiv, and Products, as well as the GraphSAGE model on Papers100M, we adopt the best-performing hyperparameter configurations reported in a recent study \cite{bajaj2024graphneuralnetworktraining}. For CoauthorCS, we run a random search over the same hyperparameter search space in that study, to identify the optimal configuration. Papers100M is too large to run a full hyperparameter search, so we select the best-performing configuration from prior work \cite{zheng2020distdgl, disthybrid, mostafa2022sequential, IBMB, salient}. For SGFormer, we use the parameters reported in the original paper~\cite{wu2023sgformer, sgformer-repo}.

We use the same hyperparameters for training on both the original and summarized graphs to ensure a fair comparison of accuracy and efficiency trade-offs.  We note that this is a \emph{conservative} choice for sparsification: the hyperparameters are tuned on the original graph and reused unchanged on the sparsified graphs, so any accuracy reported on a sparsified graph could potentially be improved by re-tuning. The gains we report are therefore lower bounds on what sparsification can achieve.

\stitle{Other training parameters.} We train all models using the cross-entropy loss function and the Adam optimizer with a learning rate of $0.001$. 
We halt the training after 50 consecutive epochs without validation improvement or after 2000 total epochs, whichever comes first; this criterion is applied consistently to both original and summarized graphs. We use mini-batch training with a batch size of 1024. We report the test accuracy of the best model, defined as the checkpoint with the highest validation accuracy, which corresponds to the model saved 50 epochs before early stopping.

\begin{figure*}[t]
\centering
\begin{subfigure}{\textwidth}
\centering
\includegraphics[width=\textwidth]{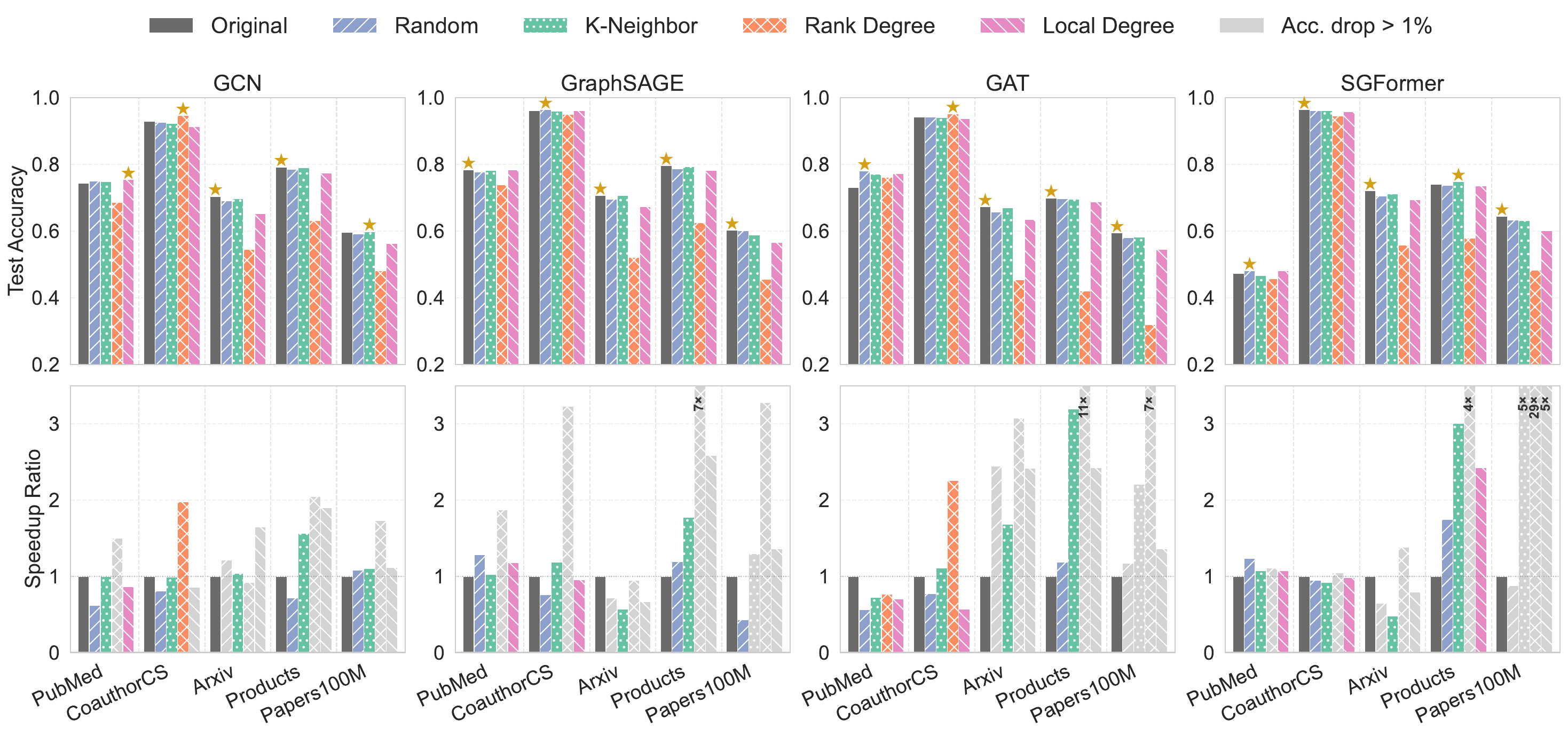}\vspace{-2mm}
\caption{Node classification}
\label{fig:accuracy_time_combined_bars_by_model}
\end{subfigure}
\vspace{2mm}
\begin{subfigure}{\textwidth}
\centering
\includegraphics[width=\textwidth]{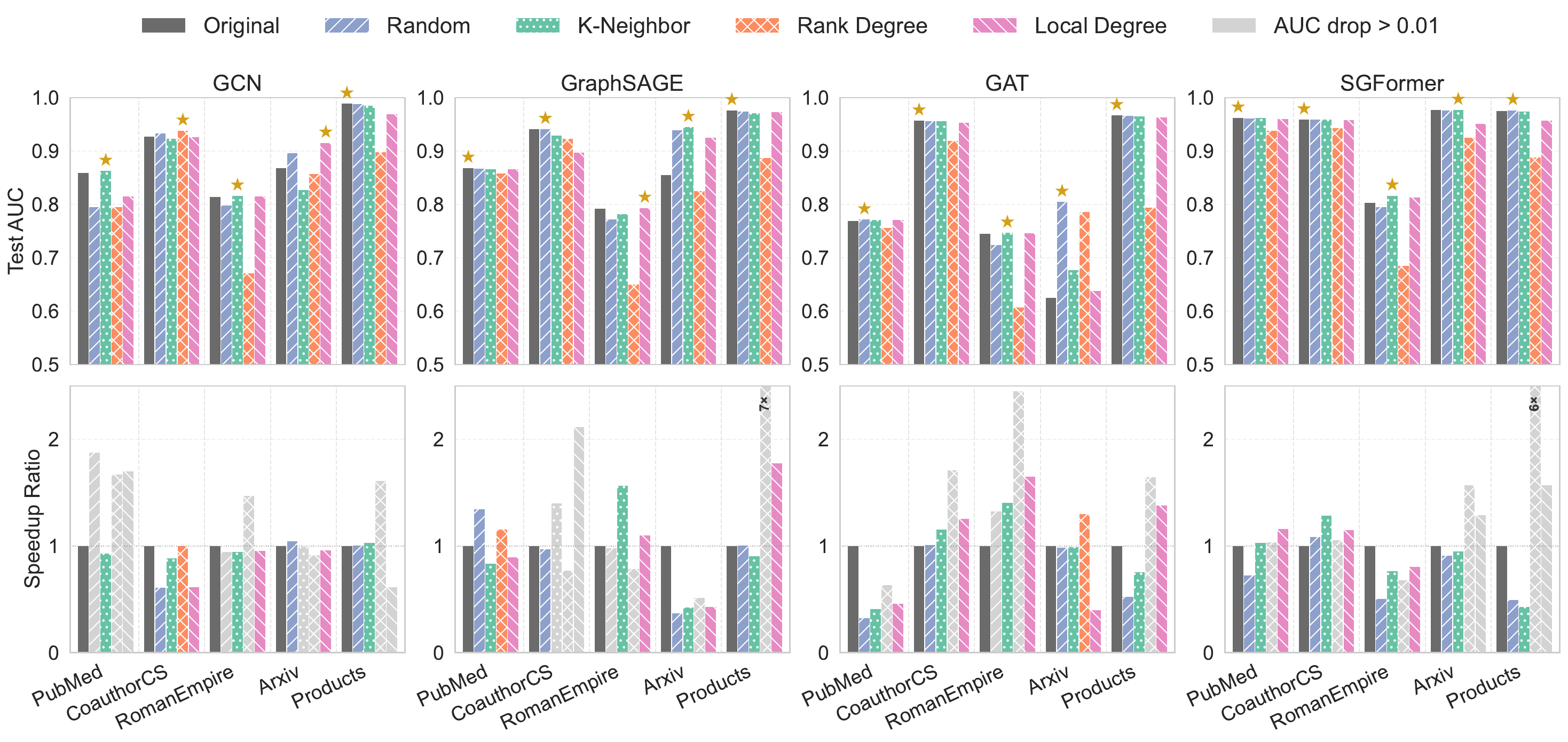}\vspace{-2mm}
\caption{Link prediction}
\label{fig:link_prediction_auc_time_bars_by_model}
\end{subfigure}
\vspace{-4mm}
\caption{Comparison of test accuracy (resp. AUC) and convergence speedup ratio of models trained on sparsified graphs over models trained on the original graphs. Star markers indicate configurations that achieved the best performance for each dataset-model pair. Light gray bars in the speedup plots indicate cases where sparsification degrades performance by more than $1\%$ accuracy for node classification or $0.01$ AUC for link prediction.}\vspace{-2mm}
\label{fig:accuracy_time_combined}
\end{figure*}

\subsection{Sparsification parameters}\label{sec:spars-params}

\label{sec:summarization-parameters}

\sysname provides an easy-to-use interface that allows users to specify custom parameters and conduct parameter sweeps to identify the best-performing configurations for their specific datasets and use cases.  In our experiments, we found there is no single parameter configuration that consistently achieves the best performance across all models and metrics. However, our results also show that accuracy does not vary substantially, as long as extreme parameter values are avoided. Based on this observation,  in this study we adopt the following parameter settings to ensure a consistent basis for comparison across experiments. 

\begin{itemize}
    \item \textbf{Random Sparsifier}: We sample the stored edge list with a removal ratio of 30\%. Since an edge is removed only if both of its stored directions are dropped (\S\ref{sec:summarization}), this removes ${\sim}9\%$ of unique edges on the graphs that store both directions, and 30\% on Arxiv and Papers100M.
    \item \textbf{K-Neighbor Sparsifier}: We set $k = 5$, such that each node retains between $\min(d_i, 5)$ and $d_i$ edges to its neighbors.
    \item \textbf{Rank Degree Sparsifier}: We use training nodes as seed nodes and iteratively add neighbors from the top 50\% by degree until 25\% of the nodes are selected.
    \item \textbf{Local Degree Sparsifier}: We set $\alpha = 0.5$, such that each node retains edges to its top $\lfloor \sqrt{\operatorname{d}(i)} \rfloor$ neighbors.
\end{itemize}\vspace{1mm}

To further demonstrate the effects of sparsification parameters, in \S\ref{sec:result-parameter-sweep}, we perform (i) a parameter sweep for all methods on the Products dataset and (ii) a comparison in which we calibrate the parameters to achieve matched compression levels across methods. 

 \section{Results}\label{sec:results}
We organize our experimental evaluation around the five questions in \S\ref{subsec:eval_goals}: accuracy and time to convergence~(Q1~\S\ref{sec:q1}), training efficiency~(Q2~\S\ref{sec:q2}), serving-time trade-offs~(Q3~\S\ref{sec:q3}), pre-processing overhead~(Q4~\S\ref{sec:q4}),~and compression vs. performance (Q5~\S\ref{sec:q5}).  In \S\ref{sec:discussion}, we provide interpretations of the results and suggest practical heuristics on when sparsification may be worthwhile.  We include all experimental data and additional plots in Appendix~\ref{app:results}.

\stitle{Experimental setup.} All experiments were conducted on a server equipped with 64 CPU cores (128 threads), 256~GB of system memory, and one NVIDIA A100 GPU with 80~GB of GPU memory. We used DGL~2.4.0 and PyTorch~2.4.0 with CUDA~12.4 and NVIDIA driver version 535.247.01.

\subsection{Accuracy and time to convergence}\label{sec:q1}

This experiment addresses \textbf{[Q1]} and aims to identify any model quality loss or gain caused by training on sparsified graphs, and to measure the corresponding effect on convergence time. 

We train the models on the original and summarized graphs produced by each of the sparsifiers, using the default parameters described in \S\ref{sec:summarization-parameters}. 
All models use early stopping triggered when validation accuracy does not improve for 50 consecutive epochs, up to a maximum of 2000 epochs, though, in practice all runs triggered early stopping much earlier.
We report the test accuracy evaluated at the model checkpoint taken 50 epochs before early stopping, along with the wall-clock time to converge.

\begin{figure*}[t]
\centering
\includegraphics[width=\textwidth]{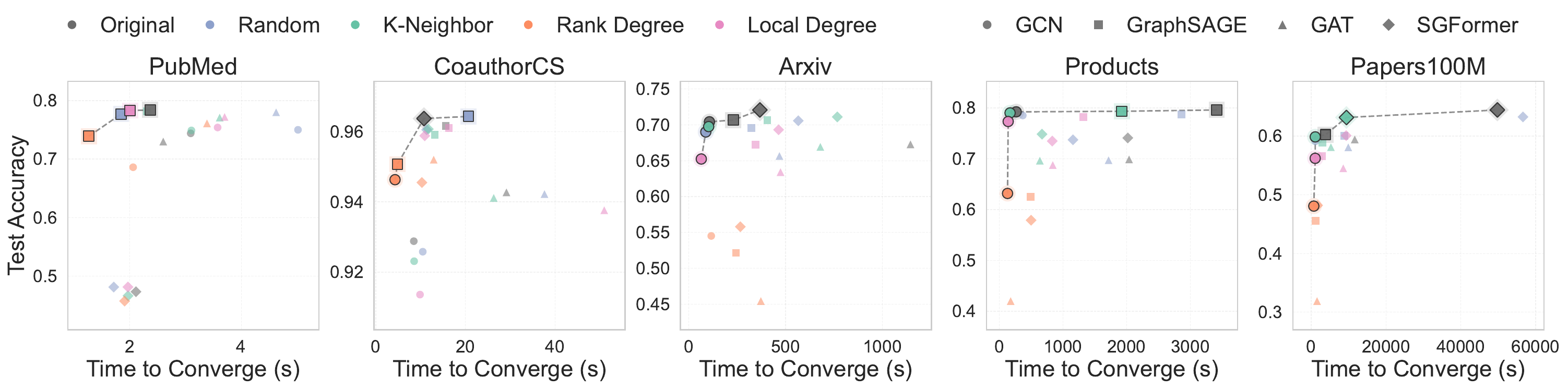}\vspace{-2mm}
\caption{Accuracy--time trade-off across graph sparsification methods. Each point shows a dataset-model-method result, with test accuracy on the y-axis and time to converge on the x-axis (in seconds). }\vspace{-2mm}
\label{fig:accuracy_vs_time_to_converge}
\end{figure*}

\begin{figure*}[t]
\centering
\includegraphics[width=\textwidth]{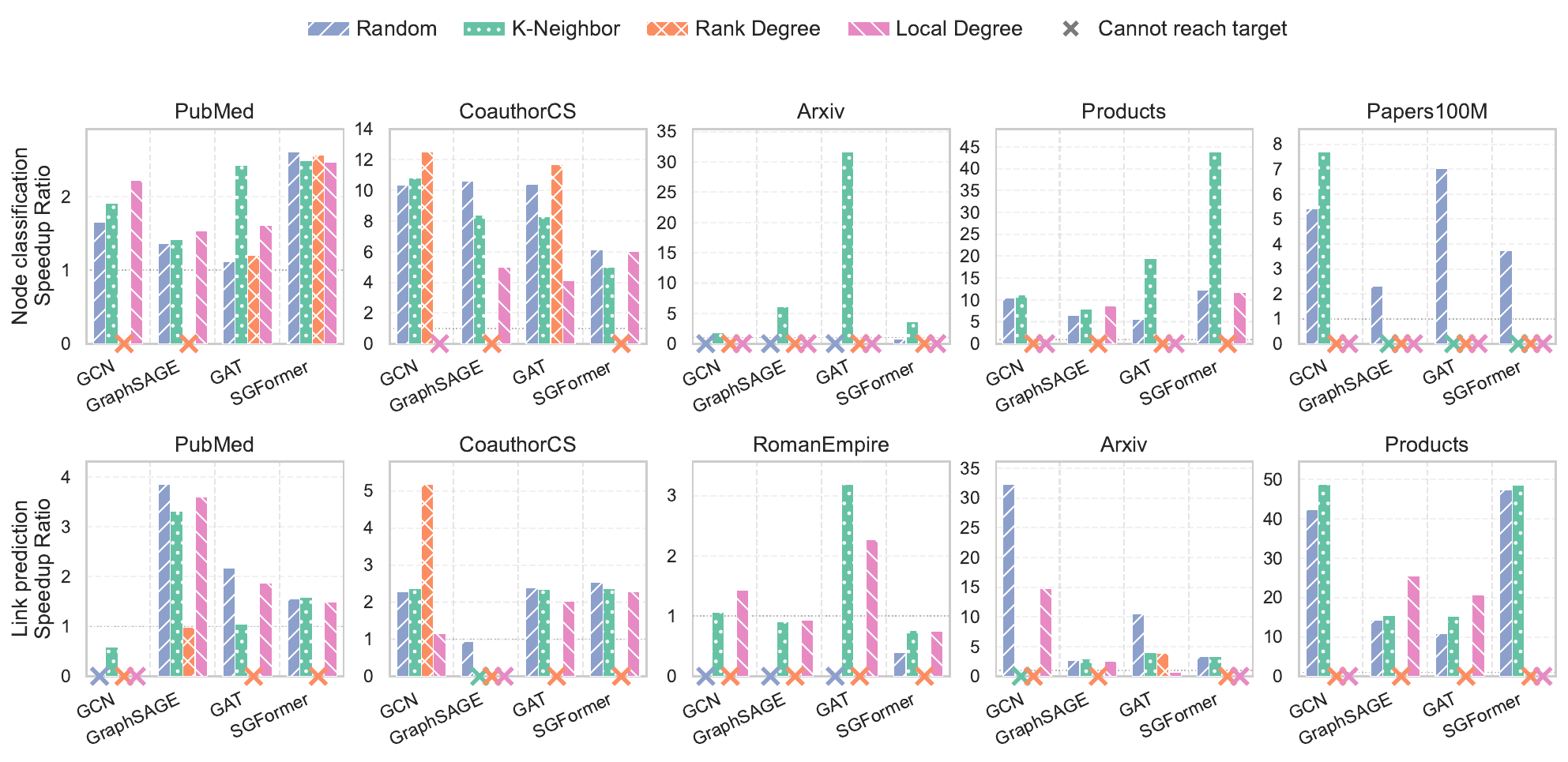}\vspace{-2mm}
\caption{Time-to-target speedup across graph sparsification methods, for node classification (top row, time-to-target \emph{accuracy}) and link prediction (bottom row, time-to-target \emph{AUC}). Each panel corresponds to a dataset, with grouped bars showing the speedup factor for each sparsification method per model. The dashed line marks speedup~$= 1\times$ (no speedup). An ``X'' marker indicates that the method could not reach the target defined by the original graph.}\vspace{-3mm}
\label{fig:time_to_accuracy_speedup_bars}
\end{figure*}

Figure~\ref{fig:accuracy_time_combined} visualizes the results as grouped bar charts for node classification and link prediction.  The top panel shows test accuracy (resp. AUC) with a star marker highlighting the best-performing method, and the bottom panel shows the speedup ratio relative to training on the original graph. 

\stitle{Test accuracy.} On the small-medium datasets (PubMed and CoauthorCS), training on sparsified graphs frequently improves accuracy over the original graph, suggesting that edge removal acts as structural regularization that reduces overfitting.
For example, on PubMed-GAT, Random sparsification raises accuracy from 0.730 to 0.780 (+6.8\%), and on CoauthorCS-GCN, Rank Degree reaches 0.946 versus 0.929 (+1.9\%).
This pattern holds across most model-method combinations on these two datasets.
On larger, sparser graphs (Arxiv, Products, Papers100M), training on the original graph achieves the best accuracy in the majority of cases. However, in all but two cases (GAT and SGFormer on Papers100M), at least one sparsification method reaches test accuracy within 1\% of the original model.

Among the four sparsifiers, K-Neighbor consistently preserves accuracy across all datasets and models, staying within 1\% of the original in most cases and even surpassing it on Papers100M-GCN (0.5986 vs.\ 0.5958). Random and Local Degree also deliver robust performance in the majority of cases. Rank Degree, in contrast, causes severe accuracy drops of 11--28 percentage points on Arxiv, Products, and Papers100M; its aggressive edge removal discards too much structural information for these larger graphs, although it can help on small, dense graphs (e.g., CoauthorCS-GCN +1.9\%, CoauthorCS-GAT +1.0\%).

Among the different models, GCN seems to benefit the most: training on a sparsified graph results in better performance than training on the original graph in three out of five datasets. SGFormer follows the same overall pattern as the message-passing models: on every dataset except Papers100M, at least one sparsification method stays within 1\% of the original graph, and on Products, K-Neighbor even surpasses it (0.7484 vs.\ 0.7410) while converging 3$\times$ faster. We note that SGFormer's absolute accuracy on PubMed is low ($\sim$0.46--0.48) regardless of sparsification, and sparsified graphs can slightly improve it (e.g., Random reaches 0.481 vs.\ 0.473 on the original graph).

The link prediction results mirror this picture. K-Neighbor and Local Degree match or exceed the full-graph AUC in 8 of 20 configurations each, while Rank Degree degrades AUC in all but two configurations. The most striking case is Arxiv, where sparsification \emph{substantially improves} link prediction: GraphSAGE rises from 0.86 to 0.95 AUC under K-Neighbor and GAT from 0.63 to 0.81 under Random. Arxiv is the one dataset where the original-graph link predictor is weak (e.g., GAT at 0.63 AUC), leaving substantial headroom; on Products the full-graph AUC is already near ceiling (0.97--0.99), so sparsification yields no gain. The Arxiv improvement is thus consistent with a regularization effect, similar to DropEdge~\cite{rong2020dropedge}, except that here it is obtained once, statically, and preserved at no recurring cost.

\begin{figure*}[t]
\centering
\includegraphics[width=\textwidth]{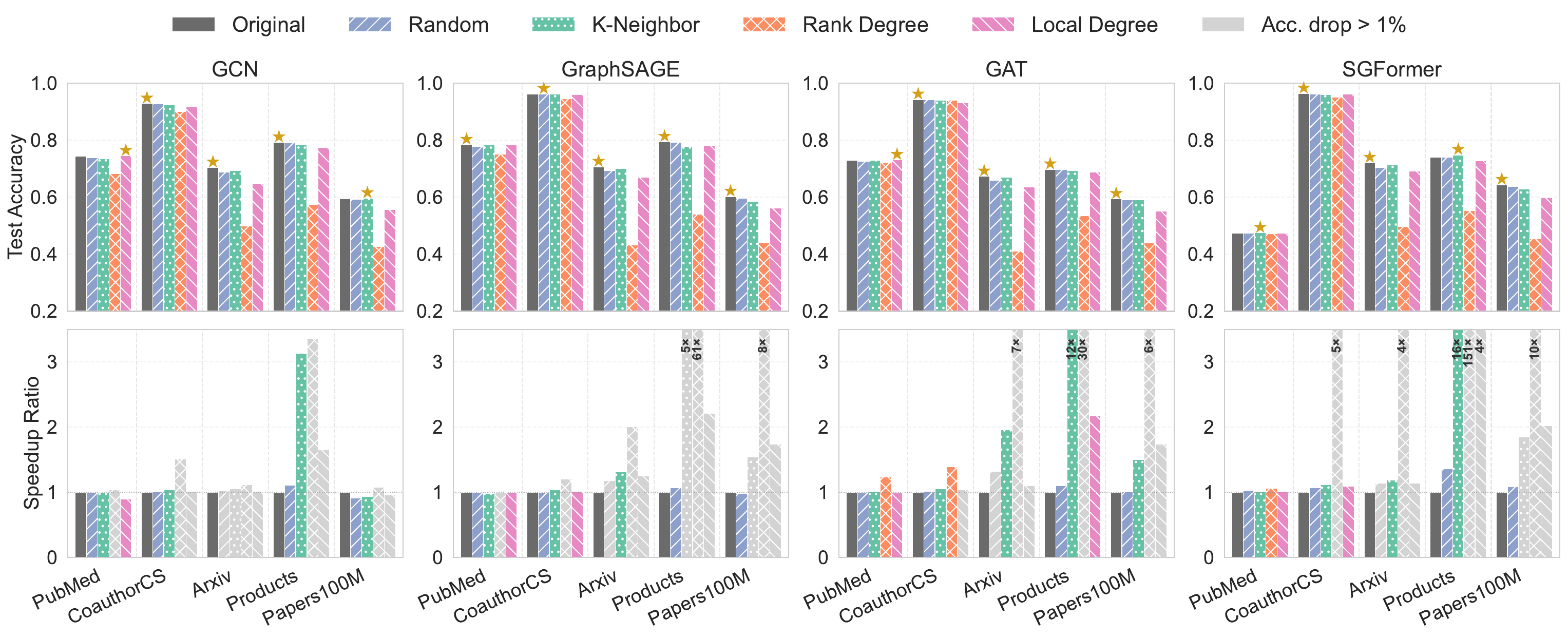}\vspace{-2mm}
\caption{Cross-graph inference comparison grouped by model. Each subplot shows one GNN architecture across all datasets. Stars mark the best method per dataset-model pair; light gray bars in the speedup plot indicate more than $1\%$ accuracy drop.}\vspace{-2mm}
\label{fig:cross_graph_inference_accuracy_time_bars_by_model}
\end{figure*}

\stitle{Speedup.} The speedup ratio plots in the bottom panels of Figure~\ref{fig:accuracy_time_combined} reveal when sparsification delivers practical benefits. Light gray bars mark cases where accuracy drops by more than 1\% (resp.\ $0.01$ for AUC) compared to the original model, indicating that the speedup comes at the cost of degraded performance. Sparsification offers minimal speedup on small datasets but substantial gains on larger ones: for example, in node classification, K-Neighbor accelerates Products-GAT convergence by 3.2$\times$ and Products-SGFormer by 3$\times$. Rank Degree's speedups, however, come with severe accuracy loss and should be interpreted with caution. Surprisingly, sparsification can also \emph{prolong} convergence: for example, Random sparsification paired with GCN on node classification slows training on all datasets except Arxiv and Papers100M, and on Arxiv, every sparsifier except Rank Degree slows down SGFormer's convergence. This demonstrates that reducing the number of edges does not guarantee shorter end-to-end training time on sparsified graphs. Note that all speedup figures in this section reflect \emph{pure training time} (neighbor sampling, feature loading, and forward/backward passes), and do \emph{not} include sparsification pre-processing overhead, measured separately in \S\ref{sec:q4}. The observed gains are driven primarily by reduced per-epoch computation---fewer edges mean smaller neighborhoods to sample and aggregate---and, on some models, by faster convergence on the sparser graph.

\stitle{Accuracy-convergence trade-offs.} \sysname can also visualize the accuracy-time trade-off as a scatter plot (Figure~\ref{fig:accuracy_vs_time_to_converge}), revealing the Pareto frontier of model-sparsifier configurations and which combinations achieve the best accuracy for a given training budget. This lets users select the optimal sparsification method for a given model and compare across models to find the configuration that best meets their accuracy and time constraints.

\noindent\underline{\textit{\textbf{Takeaways:}}}
\begin{enumerate}[leftmargin=1.2em,topsep=1pt,itemsep=1pt,parsep=0pt]
  \item Sparsification often preserves or improves accuracy / AUC.
  \item K-Neighbor is the most robust sparsification method, while Rank Degree is the least effective one.
  \item Sparsification delivers significant speedups on large but not small datasets, and can occasionally prolong training.
\end{enumerate}

\vspace{-2mm}
\subsection{Training efficiency}\label{sec:q2}

Our next experiment addresses \textbf{[Q2]} by measuring \emph{time-to-target accuracy}: how quickly a model trained on a sparsified graph matches the accuracy achieved on the original graph. We define the target as the best test accuracy on the original graph, with a 1\% tolerance margin for natural training fluctuations, and record the earliest training time at which the sparsified model reaches or exceeds it; if a method cannot reach the target before early stopping, we report a timeout. This tests a fundamentally different behavior than \S\ref{sec:q1}, as accuracy at convergence can be lower than intermediate accuracy due to overfitting. We run the same experiment for link prediction, taking each model's best test AUC on the original graph, with the same 1\% tolerance, as the target.

\stitle{Speedup.} Figure~\ref{fig:time_to_accuracy_speedup_bars} shows the speedup achieved over training on the original graph as grouped bar charts, for node classification (top row) and link prediction (bottom row). Cases marked with ``X'' could not reach the target. Note that y-axis scales differ across subplots to accommodate the range of speedups observed. We observe that speedups are most pronounced on medium-to-large graphs.
On CoauthorCS, K-Neighbor achieves 10.8$\times$ speedup for GCN and 8.4$\times$ for GraphSAGE, as the sparser graph enables faster epochs while maintaining similar convergence behavior. On Arxiv, K-Neighbor is the only method that reaches the target with a speedup, reducing training time by 1.8$\times$ on GCN, 6.1$\times$ on GraphSAGE, 3.6$\times$ on SGFormer, and exhibiting a striking 31.6$\times$ speedup on GAT (1145s $\to$ 36s); the only other method to reach a target is Random on SGFormer, which does so slower than the original graph (0.8$\times$).
On Products, Random and K-Neighbor reach the target accuracy for every model with substantial speedups: K-Neighbor achieves 11.1$\times$ for GCN, 19.5$\times$ for GAT, and 43.8$\times$ for SGFormer (2018s $\to$ 46s), while Rank Degree fails to reach the target for all models and Local Degree reaches it only for GraphSAGE (8.6$\times$) and SGFormer (11.7$\times$).
On Papers100M, only Random reaches the target for all four models, with speedups of 2.3--7.0$\times$; the only other method to reach a target is K-Neighbor on GCN (7.7$\times$).
Link prediction mirrors this pattern: gains are largest on the biggest graphs, and on Products K-Neighbor reaches the target AUC up to $49\times$ faster and Random up to $47\times$ (both on GCN and SGFormer), whereas Rank Degree fails to reach the target on all but a few small-graph configurations.

\stitle{Time-to-accuracy.} \sysname can also plot per-epoch test accuracy against cumulative training time (we show an example in Figure~\ref{fig:time_to_accuracy}, Appendix~\ref{app:results}). 
These curves are useful for detecting overfitting: if test accuracy peaks and then declines, the best accuracy occurs before the final convergence point.

\noindent\underline{\textit{\textbf{Takeaways:}}}
\begin{enumerate}[leftmargin=1.2em,topsep=1pt,itemsep=1pt,parsep=0pt]
  \item On medium/large graphs, K-Neighbor and Random can match the original graph's best accuracy with reduced training time; Rank Degree and Local Degree frequently fail to reach the target.
  \item Time-to-target accuracy amplifies the benefits: methods that preserve accuracy reach the original's performance much sooner.
\end{enumerate}

\vspace{-2mm}
\subsection{Serving-time trade-offs}\label{sec:q3}
In this section, we answer \textbf{[Q3]} by quantifying the benefits of sparsification for model serving. Specifically, we investigate whether models trained on the original graph can perform inference directly on a sparsified graph to reduce serving costs without retraining. We take the best checkpoint from training on the original graph and run forward-only inference on both the original and each sparsified graph, measuring test accuracy for fidelity loss and inference time for serving speedups. Since no retraining or fine-tuning is performed, any speedup comes \emph{for free}.

Figure~\ref{fig:cross_graph_inference_accuracy_time_bars_by_model} presents the results for node classification as grouped bar charts, organized by model.  Link prediction results are similar and given in Figure~\ref{fig:link_prediction_cross_graph_inference_auc_time_bars_by_model}, Appendix~\ref{app:results}.
Random and K-Neighbor preserve accuracy within 1--2\% of the original for most dataset-model pairs (e.g., Products-GraphSAGE: Random 0.793 vs.\ original 0.794; K-Neighbor 0.779), confirming that learned representations remain compatible with these sparsified structures. K-Neighbor also improves inference performance: on Products, it reduces GAT inference time from 413s to 35s (11.7$\times$) with only a 0.7\% accuracy drop, GraphSAGE from 838s to 181s (4.6$\times$) with a 1.9\% drop, and SGFormer from 2668s to 168s (15.9$\times$) while even slightly \emph{improving} accuracy (0.7483 vs.\ 0.7410). In contrast, Rank Degree causes severe accuracy degradation of 16--28 percentage points on Arxiv and Products (e.g., Products-GCN: 0.574 vs.\ 0.792), indicating that its aggressive structural changes are incompatible with models trained on the full graph. Overall, cross-graph inference is most effective on large datasets, where models perform costly neighborhood sampling and aggregation; on small datasets like PubMed and CoauthorCS, inference already completes within a quarter of a second regardless of sparsification, offering no practical speedup.

\noindent\underline{\textit{\textbf{Takeaways:}}}
\begin{enumerate}[leftmargin=1.2em,topsep=1pt,itemsep=1pt,parsep=0pt]
  \item K-Neighbor achieves significant inference speedups on large graphs, with at most 1--2\% accuracy loss.
  \item Cross-graph inference delivers the largest benefits on large datasets, where neighborhood sampling and aggregation dominate inference cost.
\end{enumerate}

\vspace{-2mm}
\subsection{Pre-processing overhead}\label{sec:q4}
We now evaluate the pre-processing overhead of sparsification (\textbf{[Q4]}). In practice, users pay this cost once and then train many models or architectures on the sparsified graph, so it is easily amortized.  The key practical question is when pre-processing pays off \emph{within a single training run}. Table~\ref{tab:amortization} summarizes this. Rank Degree amortizes most often (15 of 20 configurations), but recall from \S\ref{sec:q1} that its speedups come at severe accuracy loss. Among accuracy-preserving methods, K-Neighbor is most cost-effective, amortizing in 11 of 20 configurations, whereas Random performs worst at 8 of 20, as training on randomly-sparsified graphs often takes longer to converge. On the large Products dataset, nearly all method-model combinations amortize in a single run, except Random on GCN.

We further measure the end-to-end breakdown of each method's pre-processing time, including graph loading, DGL$\leftrightarrow$NumPy conversion, C++ edge-data construction and export, and the C++ sparsification routine itself (Figure~\ref{fig:summarization_time_by_dataset}, Appendix~\ref{app:results}). In brief, graph loading dominates on the small datasets while sparsification is negligible, and the C++ sparsification routine dominates on medium-large datasets. The main outlier is Rank Degree on Papers100M at 9970s ($\sim$2.8 hours), 22$\times$ slower than K-Neighbor, because the small set of training seed nodes induces many iterations.

\begin{table}[t]
\centering
\small
\setlength{\tabcolsep}{2pt}
\renewcommand{\arraystretch}{1.05}
\caption{Single-run amortization of processing costs. \checkmark{} indicates pre-processing overhead is amortized within a single training run; \xmark{} indicates its cost exceeds single-run savings. Methods per cell, left to right: R = Random, K = K-Neighbor, RD = Rank Degree, LD = Local Degree.}\vspace{-2mm}
\label{tab:amortization}
\resizebox{\columnwidth}{!}{\begin{tabular}{l cccc cccc cccc cccc}
\toprule
& \multicolumn{4}{c}{GCN} & \multicolumn{4}{c}{GraphSAGE} & \multicolumn{4}{c}{GAT} & \multicolumn{4}{c}{SGFormer} \\
\cmidrule(lr){2-5}\cmidrule(lr){6-9}\cmidrule(lr){10-13}\cmidrule(lr){14-17}
Dataset & R & K & RD & LD & R & K & RD & LD & R & K & RD & LD & R & K & RD & LD \\
\midrule
PubMed     & \xmark & \xmark & \checkmark & \xmark & \checkmark & \xmark & \checkmark & \checkmark & \xmark & \xmark & \xmark & \xmark & \checkmark & \xmark & \checkmark & \checkmark \\
CoauthorCS & \xmark & \xmark & \checkmark & \xmark & \xmark & \checkmark & \checkmark & \xmark & \xmark & \checkmark & \checkmark & \xmark & \xmark & \xmark & \checkmark & \xmark \\
Arxiv      & \checkmark & \checkmark & \xmark & \checkmark & \xmark & \xmark & \xmark & \xmark & \checkmark & \checkmark & \checkmark & \checkmark & \xmark & \xmark & \checkmark & \xmark \\
Products   & \xmark & \checkmark & \checkmark & \checkmark & \checkmark & \checkmark & \checkmark & \checkmark & \checkmark & \checkmark & \checkmark & \checkmark & \checkmark & \checkmark & \checkmark & \checkmark \\
Papers100M & \xmark & \xmark & \xmark & \xmark & \xmark & \checkmark & \xmark & \checkmark & \checkmark & \checkmark & \checkmark & \checkmark & \xmark & \checkmark & \checkmark & \checkmark \\
\bottomrule
\end{tabular}}
\end{table}

\noindent\underline{\textit{\textbf{Takeaways:}}}
\begin{enumerate}[leftmargin=1.2em,topsep=1pt,itemsep=1pt,parsep=0pt]
  \item Sparsification cost is often amortized in a single training run.
  \item As graphs scale, pre-processing cost becomes dominated by the sparsification routine.
\end{enumerate}

\begin{figure*}[t]
\centering
\includegraphics[width=\textwidth]{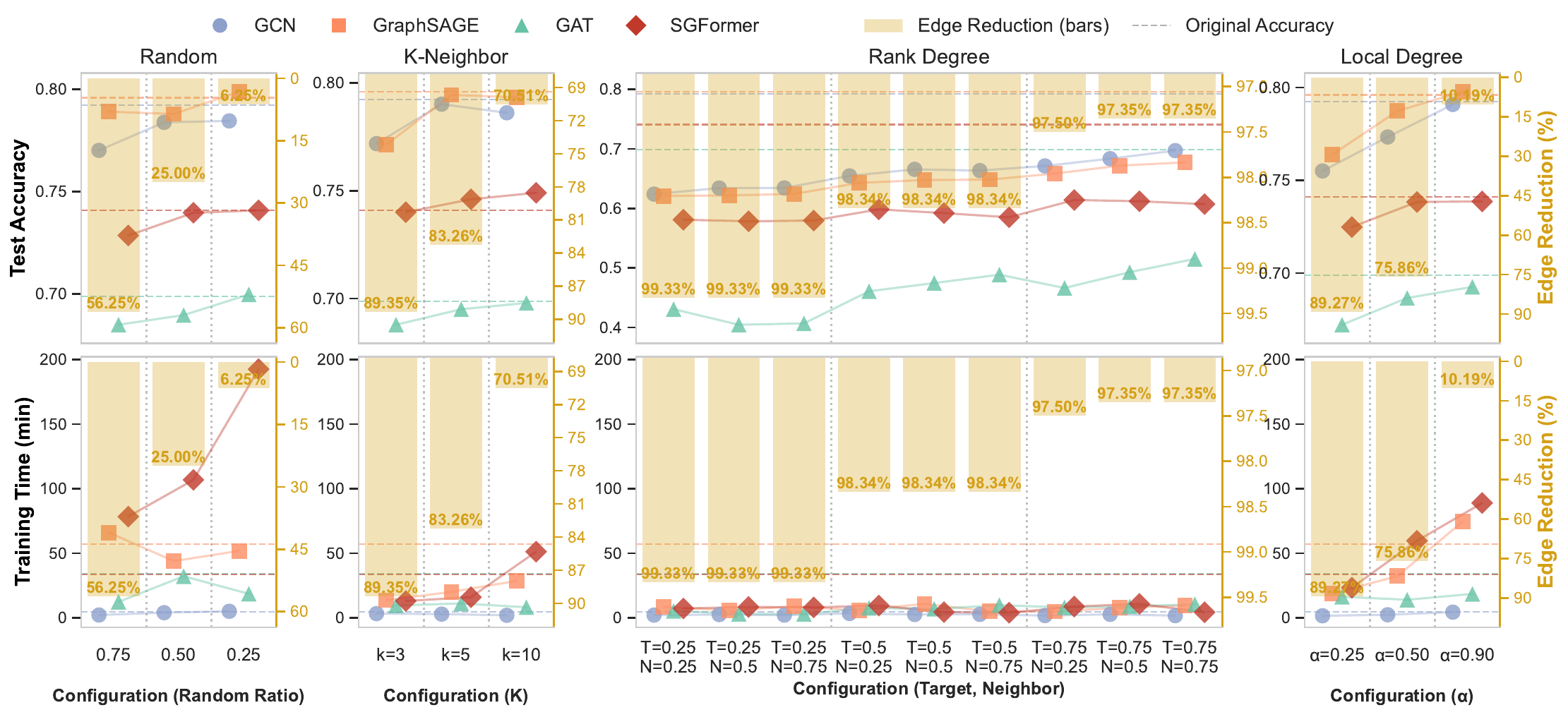}\vspace{-2mm}
\caption{Sparsification parameter sweep on the Products dataset. Each column shows one method: markers on the left axis show test accuracy (top row) and training time (bottom row) for GCN (\textcolor[HTML]{8DA0CB}{$\bullet$}), GraphSAGE (\textcolor[HTML]{FC8D62}{$\blacksquare$}), GAT (\textcolor[HTML]{66C2A5}{$\blacktriangle$}), and SGFormer (\textcolor[HTML]{c0392b}{$\blacklozenge$}); gold bars on the right axis (inverted) show the edge reduction percentage of each configuration. Dashed horizontal lines mark each model's accuracy (top) and training time (bottom) on the original graph. For Random, the axis label is the removal ratio $r$; each stored edge direction is sampled independently, the unique-edge reduction is $\approx r^2$ (\S\ref{sec:summarization}), so its gold bars fall below $r$.}\label{fig:param_sweep_accuracy_vs_reduction}
\end{figure*}
\begin{table*}[t]
\centering
\caption{Comparison of sparsifiers on Products at matched graph sizes, with hyperparameters chosen so all methods retain roughly the same fraction of unique edges (``Kept'') at three levels: L25 ($\sim$25\%), L10 ($\sim$10\%), and L$_{\min}$ ($\sim$3.8\%, the most aggressive level all methods can reach). Rank Degree appears only at L$_{\min}$ because its output is capped near a spanning forest of the selected nodes ($\sim$3.3\% retention). We report test accuracy (best value within a level marked in bold) and time to converge (s).}\vspace{-2mm}
\label{tab:fair_comparison}
\begin{tabular}{llrrrrrrrrr}
\toprule
\multirow{2}{*}{Level} & \multirow{2}{*}{Method} & \multirow{2}{*}{Kept}
& \multicolumn{2}{c}{GCN} & \multicolumn{2}{c}{GraphSAGE} & \multicolumn{2}{c}{GAT} & \multicolumn{2}{c}{SGFormer} \\
\cmidrule(lr){4-5} \cmidrule(lr){6-7} \cmidrule(lr){8-9} \cmidrule(lr){10-11}
& & & Acc. & Time (s) & Acc. & Time (s) & Acc. & Time (s) & Acc. & Time (s) \\
\midrule
\multirow{3}{*}{L25}
& Random       & 24.74\% & 0.7580 & 398 & 0.7648 & 9897 & 0.6763 & 2617 & 0.7090 & 2505 \\
& K-Neighbor   & 24.75\% & \textbf{0.7921} & 429 & \textbf{0.7950} & 8379 & \textbf{0.6958} & 1638 & \textbf{0.7499} & 1748 \\
& Local Degree & 25.05\% & 0.7750 & 738 & 0.7896 & 12406 & 0.6878 & 3637 & 0.7334 & 5077 \\
\midrule
\multirow{3}{*}{L10}
& Random       & 10.66\% & 0.7175 & 283 & 0.7386 & 5001 & 0.6487 & 2180 & 0.6866 & 1533 \\
& K-Neighbor   & 10.65\% & \textbf{0.7786} & 194 & \textbf{0.7797} & 3565 & \textbf{0.6845} & 915 & \textbf{0.7422} & 1756 \\
& Local Degree & 10.73\% & 0.7524 & 215 & 0.7650 & 4348 & 0.6682 & 2094 & 0.7215 & 1482 \\
\midrule
\multirow{4}{*}{L$_{\min}$}
& Random       & 3.80\% & 0.6631 & 205 & 0.6786 & 2187 & 0.5949 & 1001 & 0.6512 & 778 \\
& K-Neighbor   & 3.81\% & \textbf{0.7201} & 135 & \textbf{0.7214} & 715 & \textbf{0.6210} & 785 & 0.6869 & 1063 \\
& Rank Degree  & 3.30\% & 0.7028 & 215 & 0.6880 & 761 & 0.5252 & 806 & 0.6237 & 314 \\
& Local Degree & 3.87\% & 0.7143 & 137 & 0.7112 & 2951 & 0.6090 & 798 & \textbf{0.6899} & 1434 \\
\bottomrule
\end{tabular}
\end{table*}

\subsection{Compression vs. performance}\label{sec:q5}
\label{sec:result-parameter-sweep}
We address \textbf{[Q5]} with two experiments on the Products graph: a per-method parameter sweep, and a comparison at matched graph sizes isolating edge-selection quality from the resulting graph size.

\stitle{Parameter sweep.} For each method, we vary its key hyperparameters: removal ratio $\in \{0.25, 0.5, 0.75\}$ for Random,  $k \in \{3, 5, 10\}$ for K-Neighbor, target and neighbor fractions each in $\{0.25, 0.5, 0.75\}$ for Rank Degree, and $\alpha \in \{0.25, 0.5, 0.9\}$ for Local Degree. We train GCN, GraphSAGE, GAT, and SGFormer on the resulting graphs, recording edge count, edge reduction, test accuracy, and training time. Figure~\ref{fig:param_sweep_accuracy_vs_reduction} shows test accuracy (top) and training time (bottom) as parameters vary; the secondary y-axis, also annotated as inverted gold bars, gives the edge reduction percentage.

Random degrades gracefully: even its most aggressive setting (removal ratio $0.75$, removing 56\% of unique edges, cf.~\S\ref{sec:summarization}) costs only 1--3\% accuracy, making it a safe default for moderate compression. For K-Neighbor, $k{=}5$ is the sweet spot on Products, removing 83.3\% of edges while losing under 1\% accuracy for GCN (0.790 vs.\ 0.792) and GraphSAGE (0.795 vs.\ 0.796) and even improving SGFormer (0.746 vs.\ 0.741); $k{=}3$ pushes reduction to 89.4\% but drops more, and $k{=}10$ yields less compression with no consistent accuracy gain. Rank Degree is insensitive to its two hyperparameters: because it stops at a target number of \emph{vertices}, all nine configurations remove over 97.3\% of edges and collapse the graph. Local Degree tunes smoothly via $\alpha$: $\alpha{=}0.9$ removes only 10\% of edges at near-original accuracy, while $\alpha{=}0.25$ removes 89\% with 2--4\% degradation.

\stitle{Comparison at matched graph sizes.} The default parameters used in \S\ref{sec:q1} produce graphs of different sizes, so accuracy differences conflate edge-selection quality with the amount of graph retained. To decouple these, we calibrate each method to retain roughly the same fraction of unique edges at three levels: L25 ($\sim$25\%), L10 ($\sim$10\%), and L$_{\min}$ ($\sim$3.8\%), the most aggressive level every method reaches. Rank Degree is evaluated only at L$_{\min}$, as its output on Products is capped near a spanning forest of the selected nodes ($\sim$3.3\%). We train all four models per the \S\ref{sec:q1} methodology and report test accuracy and time to converge in Table~\ref{tab:fair_comparison}.

Three observations follow. First, K-Neighbor's advantage comes from its edge selection rather than retaining more edges: it achieves the best accuracy in 11 of 12 level-model combinations (Local Degree wins only for SGFormer at L$_{\min}$) and is fastest to converge in most (e.g., GraphSAGE at L10: 3565s vs.\ 4348s for Local Degree and 5001s for Random). Second, the ranking K-Neighbor $>$ Local Degree $>$ Random holds for every model and level except SGFormer at L$_{\min}$, and the gap over Random widens with sparsification (for GCN, +3.4 accuracy points at L25 but +6.1 at L10), showing that degree-aware selection matters more the fewer edges are kept; Random's competitive accuracy under default parameters (\S\ref{sec:q1}) thus reflects its modest default compression. Third, Rank Degree looks considerably better at matched sizes: at L$_{\min}$ it beats Random for GCN (0.7028 vs.\ 0.6631) and GraphSAGE (0.6880 vs.\ 0.6786) despite retaining fewer edges (3.30\% vs.\ 3.80\%), suggesting its poor \S\ref{sec:q1} showing stems from inherent extreme compression---though it remains a poor match for GAT (0.5252).

\noindent\underline{\textit{\textbf{Takeaway:}}} No single parameter configuration is best across all models and metrics, but accuracy is mostly stable as long as extreme values are avoided. At matched graph sizes, K-Neighbor remains best and Random worst, while Rank Degree's accuracy loss can be attributed to its extreme compression ceiling.

\vspace{-2mm}
\subsection{Discussion and practical guidance}\label{sec:discussion}

The experiments above establish that the four sparsifiers behave differently. Here, we offer candidate explanations for \emph{why}, relating each method's behavior to how it affects graph structure. For the original and each sparsified graph, we measure edge reduction, the shape of the degree distribution (mean and maximum degree, and the Gini coefficient of node degrees), the fraction of nodes left with no edges (a proxy for neighborhood coverage loss), the fraction of nodes in the largest connected component, the average clustering coefficient, and the edge homophily (detailed tables in Appendix~\ref{app:structural}). The following discussion is our \emph{interpretation}, offered as working hypotheses to guide future research.  To our knowledge, this is the first analysis of these effects across methods and scales.

Our accuracy results seem to depend on two graph properties, and a third could explain the speedups. The first is whether a node still has edges after sparsification. K-Neighbor and Local Degree always keep at least one edge per node, so they never isolate a node, and their accuracy stays close to the full graph. Random drops some low-degree nodes entirely (4.5--5\% on PubMed and Arxiv), matching its slightly lower accuracy. Rank Degree keeps edges only among a fixed number of chosen nodes, leaving 75--81\% of nodes with no edges, and its accuracy falls by 11--28 percentage points on the larger graphs. The second is homophily: the accuracy-preserving methods also leave homophily almost unchanged (e.g., Arxiv from 0.65 to 0.64), suggesting the edges they remove were not helping pass messages between same-label nodes, which would explain why accuracy holds or improves. The third property concerns the magnitude of compression and depends mostly on the degree distribution. K-Neighbor removes 83.3\% of edges on Products and 53.4\% on Arxiv; we believe this is why the largest speedups appear on the largest, most unevenly degreed graphs.

These observations suggest a few practical heuristics. Sparsification is most worthwhile on large, hub-dominated (heavy-tailed) graphs, where both achievable compression and speedup are largest. Methods with a per-node connectivity guarantee (K-Neighbor, Local Degree) are a safe default, whereas vertex-budget methods (Rank Degree) are best avoided when every node must be labeled. Finally, expect accuracy to be preserved or improved on homophilous graphs, and little change either way on heterophilous or near-regular ones. Validating these hypotheses across more architectures and tasks is a natural next step this study leaves open. \section{Related work}\label{sec:related}

\stitle{Graph reduction techniques for GNN tasks.}
Summarization as a pre-processing step for GNN scaling remains underexplored.  A few works demonstrate the promise of sparsification but in narrow settings.  Kosman et al.~\cite{kosman2022lsp} introduce locality-sensitive pruning to accelerate training and inference, Vatter et al.~\cite{vatter2024size} study the effects of random sparsification on sampling-based GNN training, and Kiouche et al.~\cite{kiouche2024neighborhood} propose a neighborhood-preserving method. \sysname generalizes these efforts with a holistic, extensible framework that enables the first systematic comparison of sparsifiers under a common pipeline and can readily incorporate future methods,  models, and datasets.

Beyond sparsification, prior work explores other graph reduction approaches for GNN workloads. Graph \emph{sampling}~\cite{waleffe2023mariusgnn, gong2023gsampler, smartssdsampler} extracts subgraphs during training, but repeating this across mini-batches and epochs is costly at scale. Graph \emph{coarsening}~\cite{huang2021scaling, macgnn} clusters and aggregates nodes, but alters node-level representations. Graph \emph{condensation}~\cite{exgcn, kidd2023, jin2022graph} learns a smaller synthetic graph mimicking the original, e.g., GCond~\cite{jin2022graph} matches gradients; while highly compact, it replaces the real topology with artificial structure and often needs expensive optimization. Other methods go further, with NeuralSparse~\cite{zheng2020robust} predicting unimportant edges via a separate network and inference-oriented methods~\cite{compression_inference} merging inference-equivalent nodes into quotient graphs. In contrast, \sysname uses lightweight sparsification applied as a one-time pre-processing step, amortizing its cost across epochs.

\stitle{Structural modification during and after training.}
Graph structure can also be modified dynamically during training. DropEdge~\cite{rong2020dropedge} randomly removes edges at each epoch to mitigate over-smoothing and improve generalization. Unlike static pre-processing, it is applied at every epoch, which adds repeated runtime overhead.
Other works reduce cost later in the pipeline. Liu et al.~\cite{liu2023comprehensive} propose gradual pruning of graph structure and model parameters during training. For inference-time optimization, Si et al.~\cite{si2022serving} compress the training graph by introducing virtual nodes to reduce memory lookup overhead. While effective, these approaches require modifying the core training or inference pipeline.

\stitle{Systems for scalable GNN training.}
Accelerating GNN pipelines through systems and hardware optimizations is a vibrant research area. Recent surveys~\cite{zhang2023survey, liao2025advances} and empirical evaluations~\cite{yuan1comprehensive} highlight distributed training~\cite{aineutrontp, bytegnn}, multi-GPU execution pipelines~\cite{daha, liu2023bgl}, and out-of-core storage frameworks~\cite{sheng2024outre, diskgnn, waleffe2023mariusgnn, surveyoocgnn} as key strategies for handling billion-scale graphs. 
Instead of altering the graph data, these works focus on optimizing memory access patterns, data layout, and execution planning to improve performance.

 \vspace{-2mm}
\section{Conclusion}\label{sec:conclusion}
\balance

We presented \sysname, an extensible framework that lets users seamlessly integrate graph sparsification into GNN pipelines and systematically evaluate its effects. Using it, we conducted a study spanning four sparsification methods, four GNN architectures, and six real-world graphs. Our results reveal that sparsification can preserve, and sometimes even improve, model accuracy. We relate these outcomes to a small number of structural properties and show that sparsification's pre-processing overhead is amortizable. While this paper focuses on edge reduction, exploring summarization methods that also consider nodes and features~\cite{kalavri2016shortest,wan2023adaptive} is an important direction for future work.

\ifarxiv
\begin{acks}
This work was supported by the \grantsponsor{NSF}{National Science Foundation}{https://www.nsf.gov/} under Grant No.~\href{https://www.nsf.gov/awardsearch/showAward?AWD_ID=2237193}{\grantnum{NSF}{2237193}}.
\end{acks}
\fi

\bibliographystyle{ACM-Reference-Format}
\bibliography{references}

\appendix
\onecolumn
\raggedbottom
\setlength{\intextsep}{8pt plus 2pt minus 2pt}
\setlength{\textfloatsep}{8pt plus 2pt minus 2pt}
\setlength{\floatsep}{8pt plus 2pt minus 2pt}

\section{Sparsification Algorithms}\label{app:pseudocode}

Algorithms~\ref{alg:random_sparsifier}--\ref{alg:local_degree_sparsifier} provide the pseudocode for the four sparsification methods described in \S\ref{sec:summarization}.

\begin{algorithm}[H]
\caption{Random Sparsifier}
\label{alg:random_sparsifier}
\begin{algorithmic}[1]
\Require Graph $G = (V, E)$ as an \textsl{edge list}, retention ratio $p \in (0,1]$
\Ensure Sparsified graph $G' = (V, E')$

\State Initialize $E' \leftarrow \emptyset$
\For{each edge $e \in E$}
    \State Sample $u \sim \text{Uniform}(0,1)$
    \If{$u \leq p$}
        \State $E' \leftarrow E' \cup \{e\}$
    \EndIf
\EndFor
\State \Return $G' = (V, E')$
\end{algorithmic}
\end{algorithm}

\begin{algorithm}[H]
\caption{K-Neighbor Sparsifier}
\label{alg:k_neighbor_sparsifier}
\begin{algorithmic}[1]
\Require Graph $G = (V, E)$ as \textsl{adjacency lists}, parameter $k \in \mathbb{N}$
\Ensure Sparsified graph $G' = (V, E')$

\State Initialize $E' \leftarrow \emptyset$
\For{each vertex $i \in V$}
    \State $N(i) \leftarrow \{ j \mid (i,j) \in E \}$, $d_i \leftarrow |N(i)|$
    \If{$d_i \le k$}
        \State $E' \leftarrow E' \cup \{(i,j) \mid j \in N(i)\}$
    \Else
        \State Sample a subset $S \subseteq N(i)$ of size $k$ uniformly at random
        \State $E' \leftarrow E' \cup \{(i,j) \mid j \in S\}$
    \EndIf
\EndFor
\State \Return $G' = (V, E')$
\end{algorithmic}
\end{algorithm}

\begin{algorithm}[H]
\caption{Rank Degree Sparsifier}
\label{alg:rank_degree_sparsifier}
\begin{algorithmic}[1]
\Require Graph $G=(V,E)$ as \textsl{adjacency lists}, initial seed nodes $S$, top-ratio $\rho \in (0,1]$, target sample size $x$ (no. of vertices)
\Ensure Sparsified graph $G'=(V,E')$

\State Initialize $E' \leftarrow \emptyset$
\State $\mathit{Seeds} \leftarrow$ $S$
\While{$|\mathit{V}(G')| < x$}
    \State $\mathit{SelectedEdges} \leftarrow \emptyset$
    \State $\mathit{NewSeeds} \leftarrow \emptyset$
    \ForAll{$i \in \mathit{Seeds}$}
        \State $N(i) \leftarrow \{j \in V \mid (i,j)\in E\}$
        \State Rank nodes in $N(i)$ by degree in descending order
        \State $k \leftarrow \max\!\big(1,\lfloor \rho \cdot |N(i)| \rfloor\big)$
        \State Let $j_1, j_2, \ldots, j_k$ be the top-$k$ nodes in the ranking
        \State $\mathit{SelectedEdges} \leftarrow \mathit{SelectedEdges} \cup \{(i,j_1),\ldots,(i,j_k)\} \cup \{(j_1,i),\ldots,(j_k,i)\}$
        \State $\mathit{NewSeeds} \leftarrow \mathit{NewSeeds} \cup \{j_1, j_2, \ldots, j_k\}$
    \EndFor
    \State $E' \leftarrow E' \cup \mathit{SelectedEdges}$
    \State $\mathit{Seeds} \leftarrow \mathit{NewSeeds}$
    \If{$\mathit{Seeds} = \emptyset$}
        \State $\mathit{Seeds} \leftarrow$ select $|S|$ nodes uniformly at random from $V$
    \EndIf
\EndWhile
\State \Return $G'=(V,E')$
\end{algorithmic}
\end{algorithm}

\begin{algorithm}[H]
\caption{Local Degree Sparsifier}
\label{alg:local_degree_sparsifier}
\begin{algorithmic}[1]
\Require Graph $G=(V,E)$ as \textsl{adjacency lists}, parameter $\alpha \in [0,1]$
\Ensure Sparsified graph $G'=(V,E')$

\State Initialize $E' \leftarrow \emptyset$
\For{each vertex $i \in V$}
    \State $N(i) \leftarrow \{j \mid (i,j)\in E\}$, \quad $d_i \leftarrow |N(i)|$
    \If{$d_i \not= 0$}
        \State $k \leftarrow \max\!\big(1,\lfloor \left( d_i \right)^{\alpha}\rfloor\big)$
        \State Sort $N(i)$ in descending order of degree $d_j$
        \State Let $T(i)$ be the first $k$ nodes in the sorted list
        \ForAll{$j \in T(i)$}
            \State $E' \leftarrow E' \cup \{(i,j)\}$
        \EndFor
    \EndIf
\EndFor
\State \Return $G'=(V,E')$
\end{algorithmic}
\end{algorithm}

\FloatBarrier

\section{Model Hyperparameters}\label{app:hyperparameters}

Table~\ref{tab:hyperparameters} lists the hyperparameter configurations we use for each model and dataset, selected as described in \S\ref{sec:models}. We use the same configurations for training on both the original and the sparsified graphs.

\begin{table}[H]
\centering
\caption{Model hyperparameters for different datasets.}
\renewcommand{\arraystretch}{1.05}
\begin{tabular}{llcccccc}
\toprule
\textbf{Model} & \textbf{Architecture} & \textbf{PubMed} & \textbf{CoauthorCS} & \textbf{Roman-emp.} & \textbf{Arxiv} & \textbf{Products} & \textbf{Papers100M} \\
\midrule

\multirow{3}{*}{\textbf{GraphSAGE}}
& Layers      & 2   & 2  & 2  & 5   & 5   & 3     \\
& Hidden size & 256 & 64 & 64 & 128 & 256 & 256   \\
& Fanout      & 10  & 4  & 10 & 20  & 5   & [15, 10, 5]   \\
\midrule

\multirow{4}{*}{\textbf{GAT}}
& Layers      & 2    & 2  & 2  & 4    & 3    & 3     \\
& Hidden size & 1024 & 64 & 64 & 256  & 128  & 256     \\
& Num heads   & 4    & 4  & 4  & 2    & 2    & 4     \\
& Fanout      & 10   & 20 & 10 & 10   & 5    & [15, 10, 5]      \\
\midrule

\multirow{3}{*}{\textbf{GCN}}
& Layers      & 6   & 2   & 2  & 2    & 2    & 2      \\
& Hidden size & 64  & 128 & 64 & 1024 & 512  & 128      \\
& Fanout      & 10  & 15  & 10 & 15   & 5    & 5      \\
\midrule

\multirow{5}{*}{\textbf{SGFormer}}
& GNN layers   & 2   & 2   & 2   & 3   & 3   & 3 \\
& Trans layers & 1   & 1   & 1   & 1   & 1   & 1 \\
& Hidden size  & 64  & 64  & 64  & 256 & 64  & 256 \\
& Graph weight & 0.8 & 0.8 & 0.5 & 0.5 & 0.5 & 0.5 \\
& Fanout      & [15, 10]  & [15, 10]  & [10, 10] & [15, 15, 10]   & [5, 5, 5]    & [15, 10, 5]      \\
\bottomrule

\end{tabular}
\label{tab:hyperparameters}
\end{table}

\FloatBarrier

\section{Additional Experimental Results}\label{app:results}

\subsection*{Node Classification: Accuracy and Time to Converge}

Table~\ref{tab:summarization_accuracy_time_to_converge} reports the raw test accuracy and time-to-converge measurements that we plot in Figure~\ref{fig:accuracy_time_combined_bars_by_model}.

\begin{table}[H]
\centering
\caption{Node classification: test accuracy and time to converge under different graph sparsification methods.
Time to converge is defined as the time until validation accuracy does not improve for 50 consecutive epochs.}
\label{tab:summarization_accuracy_time_to_converge}
\resizebox{\textwidth}{!}{
\begin{tabular}{llcccccccccc}
\toprule
\multirow{2}{*}{Dataset} & \multirow{2}{*}{Model}
& \multicolumn{5}{c}{Test Accuracy}
& \multicolumn{5}{c}{Time to Converge} \\
\cmidrule(lr){3-7} \cmidrule(lr){8-12}
& & Original & Random & K-Neighbor & Rank Degree & Local Degree
& Original & Random & K-Neighbor & Rank Degree & Local Degree \\
\midrule
\multirow{4}{*}{PubMed}
& GCN       & 0.7440 & 0.7500 & 0.7490 & 0.6860 & \textbf{0.7540}
& 3.0947s & 5.0175s & 3.1072s & 2.0621s & 3.5771s \\
& GraphSAGE & \textbf{0.7840} & 0.7770 & 0.7820 & 0.7390 & 0.7830
& 2.3681s & 1.8468s & 2.3080s & 1.2674s & 2.0076s \\
& GAT       & 0.7300 & \textbf{0.7800} & 0.7710 & 0.7610 & 0.7720
& 2.6020s & 4.6256s & 3.6135s & 3.3861s & 3.7024s \\
& SGFormer  & 0.4730 & \textbf{0.4810} & 0.4660 & 0.4570 & \textbf{0.4810}
& 2.1128s & 1.7146s & 1.9711s & 1.9090s & 1.9693s \\
\midrule
\multirow{4}{*}{CoauthorCS}
& GCN       & 0.9288 & 0.9258 & 0.9231 & \textbf{0.9463} & 0.9136
& 8.4755s & 10.4758s & 8.5510s & 4.2892s & 9.8619s \\
& GraphSAGE & 0.9616 & \textbf{0.9643} & 0.9591 & 0.9507 & 0.9610
& 15.6068s & 20.6720s & 13.1777s & 4.8298s & 16.3171s \\
& GAT       & 0.9427 & 0.9422 & 0.9411 & \textbf{0.9520} & 0.9376
& 29.1527s & 37.6371s & 26.3192s & 12.9100s & 50.9765s \\
& SGFormer  & \textbf{0.9637} & 0.9605 & 0.9607 & 0.9455 & 0.9588
& 10.7458s & 11.3396s & 11.7173s & 10.2904s & 10.9144s \\
\midrule
\multirow{4}{*}{Arxiv}
& GCN       & \textbf{0.7043} & 0.6901 & 0.6976 & 0.5451 & 0.6524
& 106.8146s & 87.9949s & 102.9111s & 116.1579s & 64.7462s \\
& GraphSAGE & \textbf{0.7067} & 0.6958 & 0.7065 & 0.5216 & 0.6724
& 230.1044s & 322.5018s & 405.5721s & 243.7481s & 344.9583s \\
& GAT       & \textbf{0.6728} & 0.6565 & 0.6693 & 0.4543 & 0.6341
& 1144.9218s & 468.2690s & 679.5215s & 372.4625s & 473.9342s \\
& SGFormer  & \textbf{0.7208} & 0.7057 & 0.7111 & 0.5580 & 0.6932
& 367.3889s & 565.5542s & 767.1390s & 266.4538s & 463.8810s \\
\midrule
\multirow{4}{*}{Products}
& GCN       & \textbf{0.7923} & 0.7852 & 0.7902 & 0.6318 & 0.7733
& 265.8042s & 369.2825s & 169.5967s & 129.9184s & 139.8189s \\
& GraphSAGE & \textbf{0.7960} & 0.7869 & 0.7934 & 0.6252 & 0.7820
& 3413.8999s & 2859.7327s & 1924.1599s & 489.9104s & 1321.4962s \\
& GAT       & \textbf{0.6988} & 0.6971 & 0.6962 & 0.4201 & 0.6876
& 2037.8867s & 1715.9374s & 637.4770s & 179.1285s & 840.5821s \\
& SGFormer  & 0.7410 & 0.7373 & \textbf{0.7484} & 0.5791 & 0.7348
& 2017.8683s & 1157.6282s & 671.0340s & 499.3131s & 832.2357s \\
\midrule
\multirow{4}{*}{Papers100M}
& GCN       & 0.5958 & 0.5912 & \textbf{0.5986} & 0.4807 & 0.5621
& 1174.2600s & 1084.8200s & 1068.2400s & 677.3300s & 1053.0200s \\
& GraphSAGE & \textbf{0.6023} & 0.6004 & 0.5886 & 0.4558 & 0.5659
& 3814.1200s & 8840.1800s & 2938.8500s & 1163.7200s & 2797.9900s \\
& GAT       & \textbf{0.5943} & 0.5808 & 0.5811 & 0.3189 & 0.5451
& 11625.4600s & 9898.6400s & 5261.1200s & 1553.8600s & 8561.8200s \\
& SGFormer  & \textbf{0.6445} & 0.6327 & 0.6317 & 0.4820 & 0.6008
& 49778.4894s & 56618.0211s & 9409.4981s & 1712.7204s & 9342.0695s \\
\bottomrule
\end{tabular}
}
\end{table}

\FloatBarrier

\subsection*{Link Prediction: AUC and Time to Converge}

\begin{figure}[H]
\centering
\includegraphics[width=\textwidth]{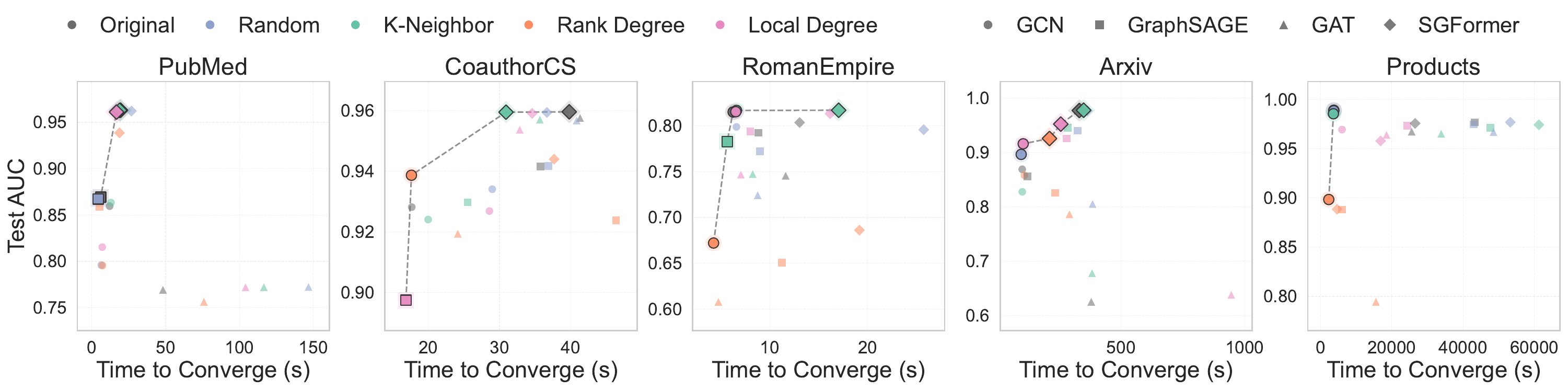}
\caption{Link prediction AUC--time trade-off across graph sparsification methods. Each point shows a dataset-model-method result, with test AUC on the y-axis and time to converge on the x-axis (in seconds).}
\label{fig:link_prediction_auc_vs_time_to_converge}
\end{figure}

Table~\ref{tab:link_prediction_auc_time_to_converge} reports the raw test AUC and time-to-converge measurements that we plot in Figure~\ref{fig:link_prediction_auc_time_bars_by_model}. Figure~\ref{fig:link_prediction_auc_vs_time_to_converge} shows the corresponding AUC--time trade-off, analogous to Figure~\ref{fig:accuracy_vs_time_to_converge} for node classification.

\begin{table}[H]
\centering
\caption{Link prediction: test AUC and time to converge under different graph sparsification methods.
Time to converge is defined as the time until validation AUC does not improve for 50 consecutive epochs.}
\label{tab:link_prediction_auc_time_to_converge}
\resizebox{\textwidth}{!}{
\begin{tabular}{llcccccccccc}
\toprule
\multirow{2}{*}{Dataset} & \multirow{2}{*}{Model}
& \multicolumn{5}{c}{Test AUC}
& \multicolumn{5}{c}{Time to Converge} \\
\cmidrule(lr){3-7} \cmidrule(lr){8-12}
& & Original & Random & K-Neighbor & Rank Degree & Local Degree
& Original & Random & K-Neighbor & Rank Degree & Local Degree \\
\midrule
\multirow{4}{*}{PubMed}
& GCN       & 0.8595 & 0.7959 & \textbf{0.8634} & 0.7955 & 0.8153
& 12.0922s & 6.4444s & 12.9958s & 7.2354s & 7.0953s \\
& GraphSAGE & \textbf{0.8691} & 0.8673 & 0.8668 & 0.8591 & 0.8669
& 6.0136s & 4.4695s & 7.2038s & 5.1967s & 6.6917s \\
& GAT       & 0.7692 & \textbf{0.7723} & 0.7721 & 0.7563 & 0.7720
& 48.1817s & 146.6801s & 116.5003s & 75.9182s & 104.0799s \\
& SGFormer  & \textbf{0.9630} & 0.9621 & 0.9629 & 0.9386 & 0.9610
& 19.4749s & 26.8557s & 18.8402s & 18.7632s & 16.7362s \\
\midrule
\multirow{4}{*}{CoauthorCS}
& GCN       & 0.9282 & 0.9341 & 0.9241 & \textbf{0.9387} & 0.9269
& 17.6860s & 28.9805s & 19.9811s & 17.6390s & 28.5806s \\
& GraphSAGE & 0.9416 & \textbf{0.9418} & 0.9298 & 0.9238 & 0.8975
& 35.7562s & 36.8455s & 25.5145s & 46.3454s & 16.8737s \\
& GAT       & \textbf{0.9576} & 0.9567 & 0.9570 & 0.9194 & 0.9537
& 41.2905s & 40.8128s & 35.6550s & 24.1443s & 32.8340s \\
& SGFormer  & \textbf{0.9596} & 0.9594 & 0.9595 & 0.9440 & 0.9591
& 39.7741s & 36.6457s & 30.9060s & 37.6522s & 34.5747s \\
\midrule
\multirow{4}{*}{Roman-empire}
& GCN       & 0.8152 & 0.7989 & \textbf{0.8169} & 0.6721 & 0.8155
& 6.2469s & 6.5848s & 6.5956s & 4.2500s & 6.5212s \\
& GraphSAGE & 0.7925 & 0.7722 & 0.7829 & 0.6508 & \textbf{0.7941}
& 8.8202s & 8.9772s & 5.6369s & 11.2398s & 8.0088s \\
& GAT       & 0.7456 & 0.7243 & \textbf{0.7474} & 0.6079 & 0.7468
& 11.6154s & 8.7602s & 8.2497s & 4.7397s & 7.0367s \\
& SGFormer  & 0.8037 & 0.7958 & \textbf{0.8171} & 0.6861 & 0.8134
& 13.0413s & 25.7219s & 17.0334s & 19.1551s & 16.1606s \\
\midrule
\multirow{4}{*}{Arxiv}
& GCN       & 0.8691 & 0.8965 & 0.8277 & 0.8581 & \textbf{0.9160}
& 99.9205s & 95.6565s & 100.1446s & 108.8104s & 103.7914s \\
& GraphSAGE & 0.8562 & 0.9401 & \textbf{0.9457} & 0.8257 & 0.9257
& 120.6950s & 323.8392s & 283.8227s & 233.2003s & 278.9049s \\
& GAT       & 0.6253 & \textbf{0.8053} & 0.6780 & 0.7863 & 0.6382
& 378.9649s & 384.4022s & 382.3147s & 290.8423s & 945.1899s \\
& SGFormer  & 0.9773 & 0.9766 & \textbf{0.9777} & 0.9257 & 0.9522
& 330.7036s & 363.5801s & 348.3135s & 210.2471s & 255.7027s \\
\midrule
\multirow{4}{*}{Products}
& GCN       & \textbf{0.9893} & 0.9888 & 0.9857 & 0.8985 & 0.9696
& 3742.3673s & 3721.9268s & 3624.3778s & 2323.3462s & 6044.8985s \\
& GraphSAGE & \textbf{0.9770} & 0.9749 & 0.9716 & 0.8882 & 0.9736
& 43305.1857s & 42977.7701s & 47695.6009s & 6058.1766s & 24333.9570s \\
& GAT       & \textbf{0.9675} & 0.9671 & 0.9653 & 0.7944 & 0.9642
& 25576.9366s & 48519.1905s & 33855.2761s & 15542.1798s & 18513.4107s \\
& SGFormer  & 0.9759 & \textbf{0.9771} & 0.9745 & 0.8886 & 0.9581
& 26545.2871s & 53229.1533s & 61243.1892s & 4652.2008s & 16864.3905s \\
\bottomrule
\end{tabular}
}
\end{table}

\FloatBarrier

\FloatBarrier

\subsection*{Time to Target Accuracy}

\begin{figure}[H]
\centering
\includegraphics[width=\textwidth]{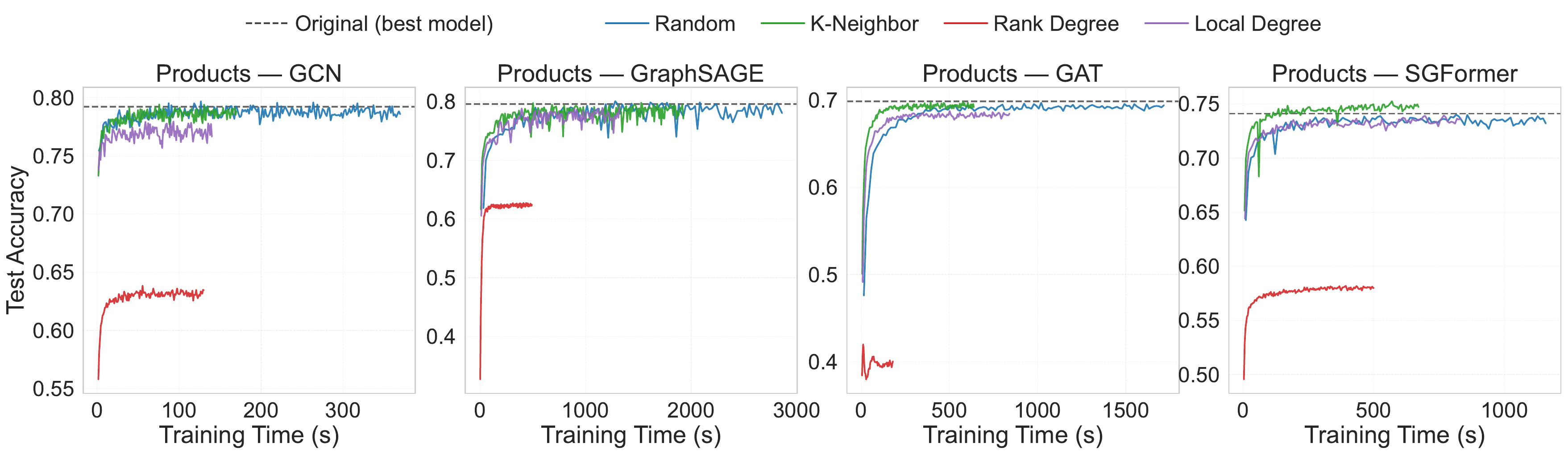}\vspace{-2mm}
\caption{Time to accuracy on the Products dataset. Each panel shows one model, with per-epoch test accuracy plotted against cumulative training time. The dashed horizontal line marks the target accuracy (within 1\% of the best test accuracy on the original graph). Sparsification methods that cross the target line sooner achieve faster time to accuracy.}\vspace{-2mm}
\label{fig:time_to_accuracy}
\end{figure}

Table~\ref{tab:time_to_accuracy_comparison} includes the raw measurements of the node classification results we plot in the top row of Figure~\ref{fig:time_to_accuracy_speedup_bars}, and Table~\ref{tab:link_prediction_time_to_accuracy_speedup} the link prediction speedups shown in its bottom row. Figure~\ref{fig:time_to_accuracy} complements them with per-epoch test accuracy curves against cumulative training time on Products.

\begin{table}[H]
\centering
\caption{Time to accuracy comparison across graph sparsification methods.
Target accuracy is defined as within 1\% of the best test accuracy achieved by the model trained on the original graph;
``--'' indicates that the method could not reach the target accuracy.}
\label{tab:time_to_accuracy_comparison}
\resizebox{\textwidth}{!}{
\begin{tabular}{llc ccccc cccc}
\toprule
\multirow{2}{*}{Dataset} & \multirow{2}{*}{Model}
& \multirow{2}{*}{Target Acc.}
& \multicolumn{5}{c}{Time to Accuracy (s)}
& \multicolumn{4}{c}{Speedup ($\times$)} \\
\cmidrule(lr){4-8} \cmidrule(lr){9-12}
& & 
& Original & Random & K-Neighbor & Rank Degree & Local Degree
& Random & K-Neighbor & Rank Degree & Local Degree \\
\midrule
\multirow{4}{*}{PubMed}
& GCN       & 0.7440 & 3.0947 & 1.8686 & 1.6193 & --     & 1.3942 & 1.6561 & 1.9112 & --      & 2.2197 \\
& GraphSAGE & 0.7840 & 2.3681 & 1.7388 & 1.6684 & --     & 1.5469 & 1.3619 & 1.4194 & --      & 1.5309 \\
& GAT       & 0.7300 & 2.6020 & 2.3309 & 1.0721 & 2.1643 & 1.6144 & 1.1163 & 2.4270 & 1.2022  & 1.6118 \\
& SGFormer & 0.4730 & 2.1128 & 0.8106 & 0.8472 & 0.8230 & 0.8557 & 2.6063 & 2.4939 & 2.5671 & 2.4691 \\
\midrule
\multirow{4}{*}{CoauthorCS}
& GCN       & 0.9288 & 8.4755  & 0.8167 & 0.7824 & 0.6776 & --     & 10.3772 & 10.8330 & 12.5089 & --     \\
& GraphSAGE & 0.9616 & 15.6068 & 1.4737 & 1.8541 & --     & 3.1321 & 10.5903 & 8.4173  & --      & 4.9829 \\
& GAT       & 0.9427 & 29.1527 & 2.7965 & 3.5158 & 2.4916 & 7.0641 & 10.4247 & 8.2920  & 11.7003 & 4.1269 \\
& SGFormer & 0.9637 & 10.7458 & 1.7583 & 2.1569 & -- & 1.7904 & 6.1115 & 4.9821 & -- & 6.0020 \\
\midrule
\multirow{4}{*}{Arxiv}
& GCN       & 0.7043 & 106.8146 & --      & 59.2351  & --      & --      & --      & 1.8032  & --       & --      \\
& GraphSAGE & 0.7067 & 230.1044 & --      & 37.8427  & --      & --      & --      & 6.0805  & --       & --      \\
& GAT       & 0.6728 & 1144.9218 & --     & 36.1945  & --      & --      & --      & 31.6324 & --       & --      \\
& SGFormer & 0.7208 & 367.3889 & 455.6653 & 101.3406 & -- & -- & 0.8063 & 3.6253 & -- & -- \\
\midrule
\multirow{4}{*}{Products}
& GCN       & 0.7923 & 265.8042  & 25.4884  & 23.8785  & --     & --      & 10.4285 & 11.1315 & --       & --      \\
& GraphSAGE & 0.7960 & 3413.8999 & 524.7011 & 426.7693 & --     & 395.7002 & 6.5064 & 7.9994  & --       & 8.6275  \\
& GAT       & 0.6988 & 2037.8867 & 364.6137 & 104.2430 & --     & --      & 5.5892 & 19.5494 & --       & --      \\
& SGFormer & 0.7410 & 2017.8683 & 164.1377 & 46.0773 & -- & 171.8308 & 12.2938 & 43.7932 & -- & 11.7433 \\
\midrule
\multirow{4}{*}{Papers100M}
& GCN       
& 0.5958 
& 1174.26 
& 216.28 
& 152.74 
& -- 
& -- 
& 5.43 
& 7.69 
& -- 
& -- \\
& GraphSAGE 
& 0.6023 
& 3814.12 
& 1648.56 
& -- 
& -- 
& -- 
& 2.31 
& -- 
& -- 
& -- \\
& GAT       
& 0.5943 
& 11625.46 
& 1652.62 
& -- 
& -- 
& -- 
& 7.03 
& -- 
& -- 
& -- \\
& SGFormer
& 0.6445
& 49778.49
& 13317.05
& --
& --
& --
& 3.74
& --
& --
& -- \\
\bottomrule
\end{tabular}
}
\end{table}

\begin{table}[H]
\centering
\caption{Link prediction: time-to-target-AUC speedup across graph sparsification methods.
The target AUC is defined as within 1\% of the best test AUC achieved by the model trained on the original graph;
``--'' indicates that the method could not reach the target AUC.}
\label{tab:link_prediction_time_to_accuracy_speedup}
\begin{tabular}{llc cccc}
\toprule
\multirow{2}{*}{Dataset} & \multirow{2}{*}{Model}
& \multirow{2}{*}{Target AUC}
& \multicolumn{4}{c}{Speedup ($\times$)} \\
\cmidrule(lr){4-7}
& &
& Random & K-Neighbor & Rank Degree & Local Degree \\
\midrule
\multirow{4}{*}{PubMed}
& GCN       & 0.8595 & -- & 0.5844 & -- & -- \\
& GraphSAGE & 0.8691 & 3.8469 & 3.3109 & 0.9871 & 3.5925 \\
& GAT       & 0.7692 & 2.1720 & 1.0555 & -- & 1.8708 \\
& SGFormer  & 0.9630 & 1.5475 & 1.5854 & -- & 1.4919 \\
\midrule
\multirow{4}{*}{CoauthorCS}
& GCN       & 0.9282 & 2.2783 & 2.3632 & 5.1660 & 1.1618 \\
& GraphSAGE & 0.9416 & 0.9286 & -- & -- & -- \\
& GAT       & 0.9576 & 2.3917 & 2.3422 & -- & 2.0266 \\
& SGFormer  & 0.9596 & 2.5301 & 2.3615 & -- & 2.2776 \\
\midrule
\multirow{4}{*}{Roman-empire}
& GCN       & 0.8152 & -- & 1.0608 & -- & 1.4361 \\
& GraphSAGE & 0.7925 & -- & 0.9113 & -- & 0.9283 \\
& GAT       & 0.7456 & -- & 3.1823 & -- & 2.2743 \\
& SGFormer  & 0.8037 & 0.3904 & 0.7667 & -- & 0.7482 \\
\midrule
\multirow{4}{*}{Arxiv}
& GCN       & 0.8691 & 32.2462 & -- & -- & 14.7271 \\
& GraphSAGE & 0.8562 & 2.6933 & 2.9798 & -- & 2.5001 \\
& GAT       & 0.6253 & 10.4989 & 4.0046 & 3.8346 & 0.6301 \\
& SGFormer  & 0.9773 & 3.3965 & 3.3890 & -- & -- \\
\midrule
\multirow{4}{*}{Products}
& GCN       & 0.9893 & 42.3797 & 48.6454 & -- & -- \\
& GraphSAGE & 0.9770 & 14.3052 & 15.3750 & -- & 25.4197 \\
& GAT       & 0.9675 & 10.8916 & 15.2714 & -- & 20.7032 \\
& SGFormer  & 0.9759 & 47.2860 & 48.4878 & -- & -- \\
\bottomrule
\end{tabular}
\end{table}

\FloatBarrier

\subsection*{Cross-Graph Inference}

\begin{figure}[H]
\centering
\includegraphics[width=\textwidth]{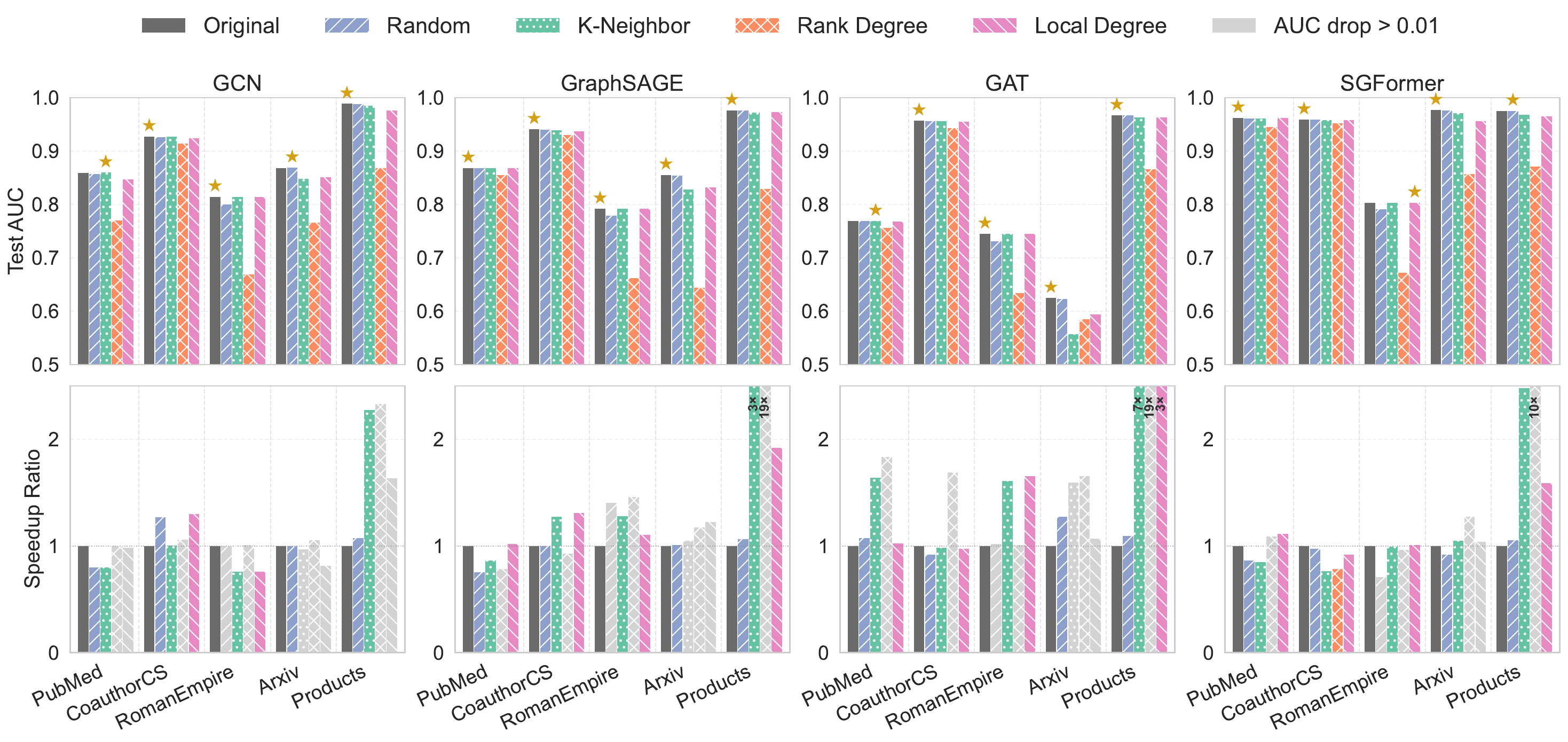}\vspace{-2mm}
\caption{Link prediction: cross-graph inference comparison grouped by model. Each subplot shows results for a single GNN architecture across all datasets. Light gray bars in the speedup plot indicate an AUC drop above $0.01$.}\vspace{-2mm}
\label{fig:link_prediction_cross_graph_inference_auc_time_bars_by_model}
\end{figure}

Table~\ref{tab:cross_graph_test_accuracy_time} reports the raw measurements behind Figure~\ref{fig:cross_graph_inference_accuracy_time_bars_by_model}. For link prediction, Figure~\ref{fig:link_prediction_cross_graph_inference_auc_time_bars_by_model} and Table~\ref{tab:link_prediction_cross_graph_auc_time} report the corresponding test AUC and inference times. K-Neighbor preserves AUC within 0.01 on PubMed, CoauthorCS, Roman-empire, and Products, while Rank Degree causes the largest degradation (up to 0.21 AUC on Arxiv-GraphSAGE). On the smaller graphs, inference already completes within a few seconds, so the resulting speedups remain below 2$\times$. On Products, however, where inference is far more expensive, sparsification yields substantial serving speedups at negligible fidelity loss: K-Neighbor reduces GAT inference time from 232s to 33s (7$\times$) with only a 0.004 AUC drop, mirroring the node-classification result.

\begin{table}[H]
\centering
\caption{Test accuracy and inference time when evaluating models trained on the original graph on both the original and sparsified graphs.}
\label{tab:cross_graph_test_accuracy_time}
\resizebox{\textwidth}{!}{
\begin{tabular}{llcccccccccc}
\toprule
\multirow{2}{*}{Dataset} & \multirow{2}{*}{Model}
& \multicolumn{5}{c}{Test Accuracy}
& \multicolumn{5}{c}{Inference Time (s)} \\
\cmidrule(lr){3-7} \cmidrule(lr){8-12}
& & Original & Random & K-Neighbor & Rank Degree & Local Degree
& Original & Random & K-Neighbor & Rank Degree & Local Degree \\
\midrule
\multirow{4}{*}{PubMed}
& GCN       & 0.7440 & 0.7390 & 0.7340 & 0.6820 & \textbf{0.7450} & 0.0206 & 0.0208 & 0.0207 & 0.0199 & 0.0230 \\
& GraphSAGE & \textbf{0.7840} & 0.7780 & 0.7830 & 0.7520 & \textbf{0.7840} & 0.0215 & 0.0216 & 0.0218 & 0.0216 & 0.0215 \\
& GAT       & 0.7300 & 0.7250 & 0.7290 & 0.7220 & \textbf{0.7310} & 0.0196 & 0.0198 & 0.0193 & 0.0159 & 0.0197 \\
& SGFormer  & 0.4730 & 0.4730 & \textbf{0.4750} & 0.4720 & 0.4740 & 0.0324 & 0.0318 & 0.0320 & 0.0306 & 0.0320 \\
\midrule
\multirow{4}{*}{CoauthorCS}
& GCN       & \textbf{0.9288} & 0.9280 & 0.9234 & 0.8997 & 0.9171 & 0.0381 & 0.0378 & 0.0368 & 0.0252 & 0.0375 \\
& GraphSAGE & 0.9616 & \textbf{0.9618} & 0.9613 & 0.9463 & 0.9599 & 0.0365 & 0.0365 & 0.0351 & 0.0303 & 0.0359 \\
& GAT       & \textbf{0.9427} & \textbf{0.9427} & 0.9406 & 0.9403 & 0.9316 & 0.0448 & 0.0440 & 0.0426 & 0.0322 & 0.0432 \\
& SGFormer  & \textbf{0.9637} & 0.9626 & 0.9597 & 0.9507 & 0.9613 & 0.2430 & 0.2274 & 0.2176 & 0.0508 & 0.2224 \\
\midrule
\multirow{4}{*}{Arxiv}
& GCN       & \textbf{0.7043} & 0.6885 & 0.6936 & 0.4996 & 0.6480 & 0.1905 & 0.1868 & 0.1811 & 0.1703 & 0.1872 \\
& GraphSAGE & \textbf{0.7067} & 0.6933 & 0.7004 & 0.4319 & 0.6697 & 0.5746 & 0.4876 & 0.4372 & 0.2858 & 0.4580 \\
& GAT       & \textbf{0.6728} & 0.6596 & 0.6700 & 0.4106 & 0.6361 & 2.4558 & 1.8581 & 1.2518 & 0.3337 & 2.2328 \\
& SGFormer  & \textbf{0.7208} & 0.7046 & 0.7127 & 0.4969 & 0.6920 & 1.6188 & 1.4152 & 1.3658 & 0.3927 & 1.4178 \\
\midrule
\multirow{4}{*}{Products}
& GCN       & \textbf{0.7923} & 0.7911 & 0.7851 & 0.5740 & 0.7742 & 25.3320 & 22.7811 & 8.0892 & 7.5313 & 15.3538 \\
& GraphSAGE & \textbf{0.7944} & 0.7928 & 0.7790 & 0.5404 & 0.7821 & 838.4105 & 783.4268 & 180.5295 & 13.8375 & 379.3458 \\
& GAT       & \textbf{0.6977} & 0.6969 & 0.6931 & 0.5356 & 0.6882 & 412.5066 & 374.7417 & 35.2860 & 13.6053 & 189.5268 \\
& SGFormer  & 0.7410 & 0.7402 & \textbf{0.7483} & 0.5535 & 0.7282 & 2668.1545 & 1965.0178 & 168.1115 & 17.6153 & 698.7332 \\
\midrule
\multirow{4}{*}{Papers100M}
& GCN
& 0.5936 & 0.5922 & \textbf{0.5965} & 0.4262 & 0.5566
& 1.1136 & 1.2202 & 1.1853 & 1.0352 & 1.1582 \\

& GraphSAGE
& \textbf{0.6021} & 0.5954 & 0.5854 & 0.4420 & 0.5619
& 8.9565 & 9.0981 & 5.8168 & 1.1863 & 5.1600 \\

& GAT
& \textbf{0.5938} & 0.5905 & 0.5911 & 0.4399 & 0.5520
& 14.8554 & 14.7069 & 9.8959 & 2.5480 & 8.5406 \\

& SGFormer
& \textbf{0.6436} & 0.6371 & 0.6283 & 0.4548 & 0.5980
& 18.2483 & 16.7649 & 9.8628 & 1.7701 & 9.0289 \\
\bottomrule
\end{tabular}
}
\end{table}

\begin{table}[H]
\centering
\caption{Link prediction: test AUC and inference time when evaluating models trained on the original graph on both the original and sparsified graphs.}
\label{tab:link_prediction_cross_graph_auc_time}
\resizebox{\textwidth}{!}{
\begin{tabular}{llcccccccccc}
\toprule
\multirow{2}{*}{Dataset} & \multirow{2}{*}{Model}
& \multicolumn{5}{c}{Test AUC}
& \multicolumn{5}{c}{Inference Time (s)} \\
\cmidrule(lr){3-7} \cmidrule(lr){8-12}
& & Original & Random & K-Neighbor & Rank Degree & Local Degree
& Original & Random & K-Neighbor & Rank Degree & Local Degree \\
\midrule
\multirow{4}{*}{PubMed}
& GCN       & 0.8595 & 0.8573 & \textbf{0.8604} & 0.7706 & 0.8473
& 0.6506 & 0.8130 & 0.8139 & 0.6482 & 0.6607 \\
& GraphSAGE & \textbf{0.8691} & 0.8690 & 0.8684 & 0.8556 & 0.8683
& 0.5196 & 0.6884 & 0.6008 & 0.6591 & 0.5078 \\
& GAT       & 0.7692 & 0.7694 & \textbf{0.7696} & 0.7564 & 0.7691
& 2.0241 & 1.8752 & 1.2329 & 1.1009 & 1.9733 \\
& SGFormer  & \textbf{0.9630} & 0.9622 & 0.9622 & 0.9455 & 0.9624
& 0.6603 & 0.7627 & 0.7747 & 0.6033 & 0.5912 \\
\midrule
\multirow{4}{*}{CoauthorCS}
& GCN       & \textbf{0.9282} & 0.9271 & 0.9273 & 0.9152 & 0.9245
& 0.7010 & 0.5505 & 0.6952 & 0.6589 & 0.5377 \\
& GraphSAGE & \textbf{0.9416} & 0.9411 & 0.9402 & 0.9307 & 0.9381
& 0.6518 & 0.6502 & 0.5112 & 0.6977 & 0.4960 \\
& GAT       & \textbf{0.9576} & 0.9568 & 0.9569 & 0.9442 & 0.9561
& 1.6708 & 1.8129 & 1.6947 & 0.9865 & 1.7121 \\
& SGFormer  & \textbf{0.9596} & 0.9593 & 0.9588 & 0.9525 & 0.9585
& 0.5939 & 0.6065 & 0.7762 & 0.7537 & 0.6430 \\
\midrule
\multirow{4}{*}{Roman-empire}
& GCN       & \textbf{0.8152} & 0.8005 & \textbf{0.8152} & 0.6699 & 0.8151
& 0.5501 & 0.5488 & 0.7222 & 0.5438 & 0.7224 \\
& GraphSAGE & \textbf{0.7925} & 0.7795 & \textbf{0.7925} & 0.6626 & \textbf{0.7925}
& 0.7584 & 0.5393 & 0.5911 & 0.5180 & 0.6863 \\
& GAT       & \textbf{0.7456} & 0.7314 & \textbf{0.7456} & 0.6351 & \textbf{0.7456}
& 1.7135 & 1.6727 & 1.0642 & 1.6947 & 1.0343 \\
& SGFormer  & 0.8037 & 0.7914 & 0.8037 & 0.6728 & \textbf{0.8038}
& 0.6109 & 0.8612 & 0.6143 & 0.6332 & 0.6045 \\
\midrule
\multirow{4}{*}{Arxiv}
& GCN       & 0.8691 & \textbf{0.8696} & 0.8492 & 0.7668 & 0.8517
& 0.7389 & 0.7373 & 0.7604 & 0.6980 & 0.9038 \\
& GraphSAGE & \textbf{0.8562} & 0.8545 & 0.8284 & 0.6443 & 0.8332
& 1.1691 & 1.1574 & 1.1106 & 0.9920 & 0.9511 \\
& GAT       & \textbf{0.6253} & 0.6238 & 0.5575 & 0.5851 & 0.5947
& 4.0714 & 3.1907 & 2.5461 & 2.4534 & 3.7976 \\
& SGFormer  & \textbf{0.9773} & 0.9765 & 0.9719 & 0.8580 & 0.9565
& 1.3944 & 1.5137 & 1.3243 & 1.0907 & 1.3375 \\
\midrule
\multirow{4}{*}{Products}
& GCN       & \textbf{0.9893} & 0.9890 & 0.9855 & 0.8685 & 0.9766
& 19.3093 & 17.9373 & 8.4854 & 8.2698 & 11.8103 \\
& GraphSAGE & \textbf{0.9770} & 0.9769 & 0.9731 & 0.8300 & 0.9740
& 215.9583 & 202.3681 & 75.4168 & 11.2172 & 112.2638 \\
& GAT       & \textbf{0.9675} & 0.9673 & 0.9639 & 0.8667 & 0.9638
& 232.3548 & 211.4870 & 33.2629 & 12.2947 & 90.7553 \\
& SGFormer  & 0.9759 & \textbf{0.9760} & 0.9688 & 0.8717 & 0.9661
& 106.9941 & 101.0773 & 43.0857 & 10.3235 & 67.1513 \\
\bottomrule
\end{tabular}
}
\end{table}

\FloatBarrier

\subsection*{Sparsification Overhead}

\begin{figure}[H]
\centering
\includegraphics[width=\textwidth]{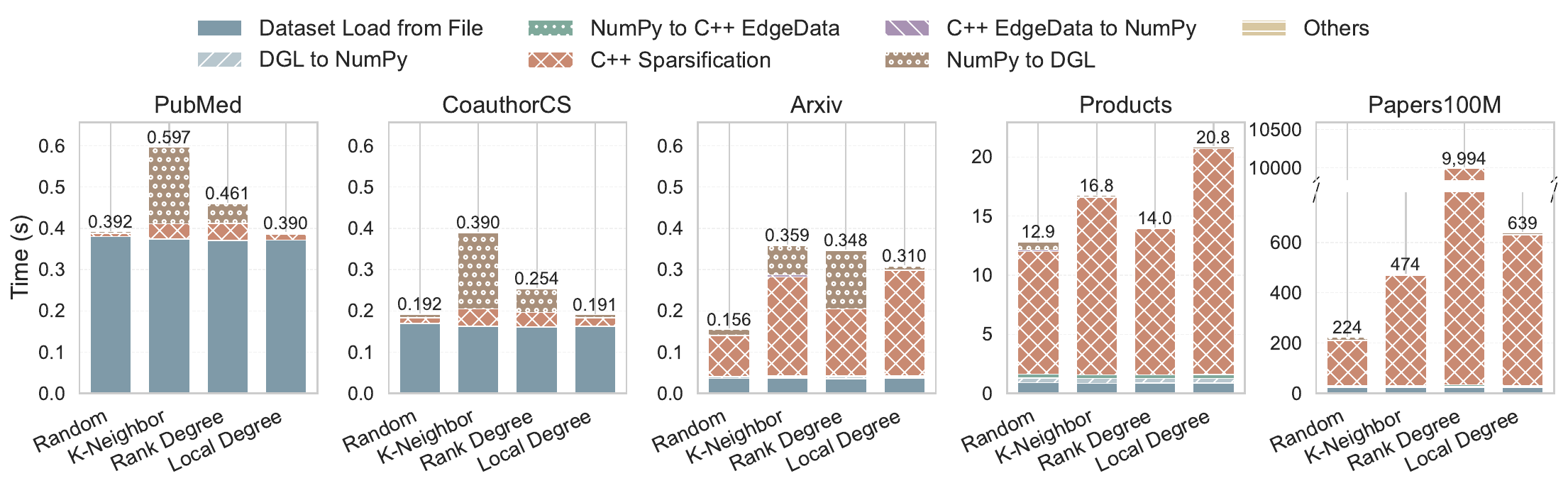}
\caption{Stacked-bar breakdown of sparsification overhead by component for each dataset. Bars include an ``Others'' segment to account for any residual time not captured by the measured components.}
\label{fig:summarization_time_by_dataset}
\end{figure}

Table~\ref{tab:summarization_time} reports the per-component overhead measurements that we aggregate in Figure~\ref{fig:summarization_time_by_dataset}, together with the edge counts and reduction ratios of each sparsified graph.
We measure the end-to-end time of each sparsification method and decompose it into the following phases:
\texttt{Dataset Load from File} (time to load the graph from file to memory),
\texttt{DGL to NumPy} (conversion from DGL to NumPy),
\texttt{NumPy to C++ EdgeData} (constructing C++ edge data from NumPy arrays),
\texttt{C++ Sparsification} (the C++ sparsification routine itself),
\texttt{C++ EdgeData to NumPy} (exporting C++ edge data back to NumPy),
and \texttt{NumPy to DGL} (conversion from NumPy back to a DGL graph).
Figure~\ref{fig:summarization_time_by_dataset} shows the results as stacked bar charts, with an ``Others'' segment for any residual time not captured by the measured components.

\begin{table}[H]
\centering
\caption{Graph sparsification overhead (s) and edge reduction statistics across datasets and sparsification methods. Edge counts are unique undirected edges. For Random, which samples each stored direction of an edge independently (\S\ref{sec:summarization}), an edge is removed only when both directed copies are removed, so its effective reduction is lower than the nominal removal ratio of $0.3$ on graphs whose loaders store both directions.}
\label{tab:summarization_time}
\scriptsize
\setlength{\tabcolsep}{3pt}
\renewcommand{\arraystretch}{1.1}
\resizebox{\textwidth}{!}{
\begin{tabular}{llrrrrrrrrrrrr}
\toprule
Dataset & Method & \#Edges (Orig.) & \#Edges (Sum.) & Reduction Ratio & Dataset Load from File & DGL to NumPy & NumPy to C++ EdgeData & C++ Sparsification & C++ EdgeData to NumPy & NumPy to DGL & C++ Total & Total \\
\midrule
\multirow{4}{*}{PubMed}
& Random & 44324 & 40298 & 0.0908 & 0.3797 & 0.0003 & 0.0001 & 0.0078 & 0.0003 & 0.0034 & 0.0082 & 0.0126 \\
& K-Neighbor & 44324 & 36209 & 0.1831 & 0.3743 & 0.0003 & 0.0001 & 0.0364 & 0.0003 & 0.1851 & 0.0369 & 0.2231 \\
& Rank Degree & 44324 & 4870 & 0.8901 & 0.3705 & 0.0004 & 0.0001 & 0.0410 & 0.0000 & 0.0476 & 0.0411 & 0.0900 \\
& Local Degree & 44324 & 35866 & 0.1908 & 0.3715 & 0.0003 & 0.0001 & 0.0141 & 0.0001 & 0.0027 & 0.0144 & 0.0181 \\
\midrule
\multirow{4}{*}{CoauthorCS}
& Random & 81894 & 74542 & 0.0898 & 0.1686 & 0.0005 & 0.0003 & 0.0140 & 0.0005 & 0.0077 & 0.0148 & 0.0237 \\
& K-Neighbor & 81894 & 57136 & 0.3023 & 0.1623 & 0.0005 & 0.0003 & 0.0416 & 0.0006 & 0.1834 & 0.0426 & 0.2274 \\
& Rank Degree & 81894 & 4815 & 0.9412 & 0.1602 & 0.0006 & 0.0003 & 0.0350 & 0.0001 & 0.0565 & 0.0354 & 0.0937 \\
& Local Degree & 81894 & 53558 & 0.3460 & 0.1623 & 0.0005 & 0.0004 & 0.0198 & 0.0002 & 0.0076 & 0.0205 & 0.0291 \\
\midrule
\multirow{4}{*}{Arxiv}
& Random & 1157799 & 813286 & 0.2976 & 0.0359 & 0.0032 & 0.0026 & 0.0977 & 0.0023 & 0.0135 & 0.1026 & 0.1198 \\
& K-Neighbor & 1157799 & 539158 & 0.5343 & 0.0371 & 0.0036 & 0.0025 & 0.2393 & 0.0045 & 0.0705 & 0.2464 & 0.3219 \\
& Rank Degree & 1157799 & 81469 & 0.9296 & 0.0353 & 0.0038 & 0.0030 & 0.1629 & 0.0004 & 0.1411 & 0.1663 & 0.3123 \\
& Local Degree & 1157799 & 536995 & 0.5362 & 0.0370 & 0.0034 & 0.0028 & 0.2542 & 0.0015 & 0.0099 & 0.2585 & 0.2729 \\
\midrule
\multirow{4}{*}{Products}
& Random & 61859012 & 56293478 & 0.0900 & 0.9250 & 0.3745 & 0.3376 & 10.3577 & 0.1471 & 0.6746 & 10.8424 & 11.9378 \\
& K-Neighbor & 61859012 & 10355844 & 0.8326 & 0.8659 & 0.3794 & 0.3401 & 15.0112 & 0.0293 & 0.1099 & 15.3806 & 15.8991 \\
& Rank Degree & 61859012 & 415643 & 0.9933 & 0.8704 & 0.3757 & 0.3383 & 12.3187 & 0.0012 & 0.0310 & 12.6583 & 13.0895 \\
& Local Degree & 61859012 & 14934280 & 0.7586 & 0.8755 & 0.3757 & 0.3398 & 19.0911 & 0.0287 & 0.1058 & 19.4596 & 19.9695 \\
\midrule
\multirow{4}{*}{Papers100M}
& Random & 1614062356 & 1130183170 & 0.2998 & 22.3113 & 4.6314 & 4.3350 & 178.7465 & 2.3682 & 11.3174 & 185.4497 & 201.7513 \\
& K-Neighbor & 1614062356 & 399060868 & 0.7528 & 22.3077 & 4.6260 & 4.3438 & 438.2324 & 1.1956 & 3.2746 & 443.7718 & 451.9862 \\
& Rank Degree & 1614062356 & 26557810 & 0.9835 & 24.2354 & 4.6285 & 4.3522 & 9960.5594 & 0.1402 & 0.2562 & 9965.0518 & 9970.1783 \\
& Local Degree & 1614062356 & 501741847 & 0.6891 & 22.7608 & 4.6089 & 4.3312 & 598.4115 & 2.5198 & 5.9952 & 605.2624 & 616.2161 \\
\bottomrule
\end{tabular}
}
\end{table}

\FloatBarrier

\subsection*{Products Parameter Sweep}

Tables~\ref{tab:products_param_sweep_gcn}--\ref{tab:products_param_sweep_sgformer} report the per-configuration measurements of the parameter sweep that we summarize in Figure~\ref{fig:param_sweep_accuracy_vs_reduction}, with one table per model.

\begin{table}[H]
\centering
\caption{\textbf{GCN} on Products dataset parameter sweep results. Edge counts are unique undirected edges. Random sampling operates on the edge list that stores both directions of each edge, and an edge is removed only if both of its directions are removed, so its effective removal is lower than the nominal ratio (\S\ref{sec:summarization}).}
\label{tab:products_param_sweep_gcn}
\begin{tabular}{llrrrr}
\toprule
Method & Config & \multicolumn{1}{c}{\#Edges} & \multicolumn{1}{c}{Reduction (\%)} & \multicolumn{1}{c}{Test Accuracy} & \multicolumn{1}{c}{Train Time (s)} \\
\midrule
Original
& baseline & 61859012 & 0.00\% & 0.7923 & 265.8042 \\
\midrule
\multirow{3}{*}{Random}
& ratio=0.25 & 57993948 & 6.25\% & 0.7846 & 299.3513\\
& ratio=0.5  & 46392065 & 25.00\% & 0.7839 & 235.6120 \\
& ratio=0.75 & 27064456 & 56.25\% & 0.7702 & 127.7833 \\
\midrule
\multirow{3}{*}{K-Neighbor}
& k=3  & 6590411 & 89.35\% & 0.7719 & 191.7707 \\
& k=5  & 10355844 & 83.26\% & 0.7902 & 168.1629 \\
& k=10 & 18242324 & 70.51\% & 0.7862 & 114.0947 \\
\midrule
\multirow{9}{*}{Rank Degree}
& target=0.25, neighbor=0.25 & 415643  & 99.33\% & 0.6240 & 122.8845 \\
& target=0.25, neighbor=0.5  & 415643  & 99.33\% & 0.6342 & 146.4399 \\
& target=0.25, neighbor=0.75 & 415643  & 99.33\% & 0.6344 & 122.5082 \\
& target=0.5,  neighbor=0.25 & 1027900 & 98.34\% & 0.6546 & 197.6375 \\
& target=0.5,  neighbor=0.5  & 1027900 & 98.34\% & 0.6656 & 151.8082 \\
& target=0.5,  neighbor=0.75 & 1027900 & 98.34\% & 0.6635 & 156.7397 \\
& target=0.75, neighbor=0.25 & 1543774 & 97.50\% & 0.6713 & 101.4903 \\
& target=0.75, neighbor=0.5  & 1640157 & 97.35\% & 0.6835 & 157.2378 \\
& target=0.75, neighbor=0.75 & 1640157 & 97.35\% & 0.6970 & 99.0456 \\
\midrule
\multirow{3}{*}{Local Degree}
& alpha=0.25 & 6635222  & 89.27\% & 0.7550 & 91.1614 \\
& alpha=0.5  & 14934280 & 75.86\% & 0.7733 & 137.7781 \\
& alpha=0.9  & 55555718 & 10.19\% & 0.7908 & 260.0897 \\
\bottomrule
\end{tabular}
\end{table}

\begin{table}[H]
\centering
\caption{\textbf{GraphSAGE} on Products dataset parameter sweep results. Edge counts are unique undirected edges. Random sampling operates on the edge list that stores both directions of each edge, and an edge is removed only if both of its directions are removed, so its effective removal is lower than the nominal ratio (\S\ref{sec:summarization}).}
\label{tab:products_param_sweep_graphsage}
\begin{tabular}{llrrrr}
\toprule
Method & Config & \multicolumn{1}{c}{\#Edges} & \multicolumn{1}{c}{Reduction (\%)} & \multicolumn{1}{c}{Test Accuracy} & \multicolumn{1}{c}{Train Time (s)} \\
\midrule
Original
& baseline & 61859012 & 0.00\% & 0.7960 & 3413.8999 \\
\midrule
\multirow{3}{*}{Random}
& ratio=0.25 & 57993948 & 6.25\% & 0.7989 & 3106.9166 \\
& ratio=0.5  & 46392065 & 25.00\% & 0.7880 & 2631.9000 \\
& ratio=0.75 & 27064456 & 56.25\% & 0.7891 & 3958.7895 \\
\midrule
\multirow{3}{*}{K-Neighbor}
& k=3  & 6590411 & 89.35\% & 0.7714 & 813.4175 \\
& k=5  & 10355844 & 83.26\% & 0.7945 & 1202.8333 \\
& k=10 & 18242324 & 70.51\% & 0.7934 & 1710.1249 \\
\midrule
\multirow{9}{*}{Rank Degree}
& target=0.25, neighbor=0.25 & 415643 & 99.33\% & 0.6209 & 508.0948 \\
& target=0.25, neighbor=0.5  & 415643 & 99.33\% & 0.6218 & 337.8445 \\
& target=0.25, neighbor=0.75 & 415643 & 99.33\% & 0.6243 & 527.2100 \\
& target=0.5,  neighbor=0.25 & 1027900 & 98.34\% & 0.6429 & 347.4103 \\
& target=0.5,  neighbor=0.5  & 1027900 & 98.34\% & 0.6476 & 642.3504 \\
& target=0.5,  neighbor=0.75 & 1027900 & 98.34\% & 0.6483 & 298.1944 \\
& target=0.75, neighbor=0.25 & 1543774 & 97.50\% & 0.6585 & 290.2142 \\
& target=0.75, neighbor=0.5  & 1640157 & 97.35\% & 0.6718 & 457.4046 \\
& target=0.75, neighbor=0.75 & 1640157 & 97.35\% & 0.6773 & 564.6107 \\
\midrule
\multirow{3}{*}{Local Degree}
& alpha=0.25 & 6635222 & 89.27\% & 0.7638 & 1123.9686 \\
& alpha=0.5  & 14934280 & 75.86\% & 0.7873 & 1935.1525 \\
& alpha=0.9  & 55555718 & 10.19\% & 0.7977 & 4468.4467 \\
\bottomrule
\end{tabular}
\end{table}

\begin{table}[H]
\centering
\caption{\textbf{GAT} on Products dataset parameter sweep results. Edge counts are unique undirected edges. Random sampling operates on the edge list that stores both directions of each edge, and an edge is removed only if both of its directions are removed, so its effective removal is lower than the nominal ratio (\S\ref{sec:summarization}).}
\label{tab:products_param_sweep_gat}
\begin{tabular}{llrrrr}
\toprule
Method & Config & \multicolumn{1}{c}{\#Edges} & \multicolumn{1}{c}{Reduction (\%)} & \multicolumn{1}{c}{Test Accuracy} & \multicolumn{1}{c}{Train Time (s)} \\
\midrule
Original
& baseline & 61859012 & 0.00\% & 0.6988 & 2037.8867 \\
\midrule
\multirow{3}{*}{Random}
& ratio=0.25 & 57993948 & 6.25\% & 0.6998 & 1113.7823 \\
& ratio=0.5  & 46392065 & 25.00\% & 0.6897 & 1926.5180 \\
& ratio=0.75 & 27064456 & 56.25\% & 0.6850 & 736.1083 \\
\midrule
\multirow{3}{*}{K-Neighbor}
& k=3  & 6590411 & 89.35\% & 0.6879 & 568.2854 \\
& k=5  & 10355844 & 83.26\% & 0.6951 & 663.4157 \\
& k=10 & 18242324 & 70.51\% & 0.6980 & 488.0483 \\
\midrule
\multirow{9}{*}{Rank Degree}
& target=0.25, neighbor=0.25 & 415643 & 99.33\% & 0.4308 & 315.7575 \\
& target=0.25, neighbor=0.5  & 415643 & 99.33\% & 0.4047 & 178.9809 \\
& target=0.25, neighbor=0.75 & 415643 & 99.33\% & 0.4069 & 172.7048 \\
& target=0.5,  neighbor=0.25 & 1027900 & 98.34\% & 0.4609 & 456.5613 \\
& target=0.5,  neighbor=0.5  & 1027900 & 98.34\% & 0.4747 & 410.7593 \\
& target=0.5,  neighbor=0.75 & 1027900 & 98.34\% & 0.4887 & 583.2249 \\
& target=0.75, neighbor=0.25 & 1543774 & 97.50\% & 0.4664 & 481.2082 \\
& target=0.75, neighbor=0.5  & 1640157 & 97.35\% & 0.4929 & 523.3901 \\
& target=0.75, neighbor=0.75 & 1640157 & 97.35\% & 0.5152 & 626.5726 \\
\midrule
\multirow{3}{*}{Local Degree}
& alpha=0.25 & 6635222 & 89.27\% & 0.6722 & 973.0815 \\
& alpha=0.5  & 14934280 & 75.86\% & 0.6866 & 835.3740 \\
& alpha=0.9  & 55555718 & 10.19\% & 0.6926 & 1106.8288 \\
\bottomrule
\end{tabular}
\end{table}

\begin{table}[H]
\centering
\caption{\textbf{SGFormer} on Products dataset parameter sweep results. Edge counts are unique undirected edges. Random sampling operates on the edge list that stores both directions of each edge, and an edge is removed only if both of its directions are removed, so its effective removal is lower than the nominal ratio (\S\ref{sec:summarization}).}
\label{tab:products_param_sweep_sgformer}
\begin{tabular}{llrrrr}
\toprule
Method & Config & \multicolumn{1}{c}{\#Edges} & \multicolumn{1}{c}{Reduction (\%)} & \multicolumn{1}{c}{Test Accuracy} & \multicolumn{1}{c}{Train Time (s)} \\
\midrule
Original
& baseline & 61859012 & 0.00\% & 0.7410 & 2017.87 \\
\midrule
\multirow{3}{*}{Random}
& ratio=0.25 & 57993948 & 6.25\% & 0.7409 & 11549.28 \\
& ratio=0.5  & 46392065 & 25.00\% & 0.7397 & 6406.34 \\
& ratio=0.75 & 27064456 & 56.25\% & 0.7286 & 4698.28 \\
\midrule
\multirow{3}{*}{K-Neighbor}
& k=3  & 6590411  & 89.35\% & 0.7401 & 777.53 \\
& k=5  & 10355844 & 83.26\% & 0.7461 & 955.32 \\
& k=10 & 18242324 & 70.51\% & 0.7491 & 3068.68 \\
\midrule
\multirow{9}{*}{Rank Degree}
& target=0.25, neighbor=0.25 & 415643  & 99.33\% & 0.5810 & 428.48 \\
& target=0.25, neighbor=0.5  & 415643  & 99.33\% & 0.5784 & 500.96 \\
& target=0.25, neighbor=0.75 & 415643  & 99.33\% & 0.5796 & 479.63 \\
& target=0.5,  neighbor=0.25 & 1027900 & 98.34\% & 0.5984 & 566.96 \\
& target=0.5,  neighbor=0.5  & 1027900 & 98.34\% & 0.5921 & 260.12 \\
& target=0.5,  neighbor=0.75 & 1027900 & 98.34\% & 0.5851 & 231.70 \\
& target=0.75, neighbor=0.25 & 1574066 & 97.46\% & 0.6141 & 519.74 \\
& target=0.75, neighbor=0.5  & 1640157 & 97.35\% & 0.6120 & 644.05 \\
& target=0.75, neighbor=0.75 & 1640157 & 97.35\% & 0.6072 & 252.49 \\
\midrule
\multirow{3}{*}{Local Degree}
& alpha=0.25 & 6635222  & 89.27\% & 0.7248 & 1396.56 \\
& alpha=0.5  & 14934280 & 75.86\% & 0.7383 & 3575.08 \\
& alpha=0.9  & 55555718 & 10.19\% & 0.7386 & 5330.81 \\
\bottomrule
\end{tabular}
\end{table}

\FloatBarrier

\section{Structural Statistics of the Sparsified Graphs}\label{app:structural}

Table~\ref{tab:sparsified_graph_stats} reports the structural statistics of the original and sparsified graphs that we use to interpret the results in \S\ref{sec:discussion}.

\begin{table}[H]
\centering
\caption{Structural statistics of the original and sparsified graphs, computed on the undirected version of each graph: mean and maximum degree, Gini coefficient of the degree distribution, fraction of isolated nodes (degree zero), fraction of nodes in the largest connected component, average clustering coefficient, and edge homophily. Connectivity and clustering statistics for Papers100M are omitted (--) due to its scale. Note that 2\% of Products nodes are already isolated in the original graph.}
\label{tab:sparsified_graph_stats}
\small
\setlength{\tabcolsep}{4pt}
\renewcommand{\arraystretch}{1.1}
\begin{tabular}{llrrrrrrr}
\toprule
Dataset & Method & Mean Deg. & Max Deg. & Deg. Gini & Isolated (\%) & Largest CC (\%) & Avg. Clust. & Edge Hom. \\
\midrule
\multirow{5}{*}{PubMed}
& Original     & 4.5 & 171 & 0.60 & 0.0  & 100.0 & 0.06 & 0.80 \\
& Random       & 4.1 & 150 & 0.61 & 4.5  & 95.2  & 0.05 & 0.80 \\
& K-Neighbor   & 3.7 & 104 & 0.54 & 0.0  & 100.0 & 0.03 & 0.80 \\
& Rank Degree  & 0.5 & 78  & 0.86 & 75.0 & 5.0   & 0.00 & 0.79 \\
& Local Degree & 3.6 & 171 & 0.57 & 0.0  & 100.0 & 0.06 & 0.80 \\
\midrule
\multirow{5}{*}{CoauthorCS}
& Original     & 8.9 & 136 & 0.46 & 0.0  & 100.0 & 0.34 & 0.81 \\
& Random       & 8.1 & 125 & 0.46 & 0.6  & 99.1  & 0.30 & 0.81 \\
& K-Neighbor   & 6.2 & 57  & 0.34 & 0.0  & 100.0 & 0.25 & 0.82 \\
& Rank Degree  & 0.5 & 24  & 0.88 & 80.9 & 15.1  & 0.04 & 0.80 \\
& Local Degree & 5.8 & 136 & 0.49 & 0.0  & 100.0 & 0.32 & 0.80 \\
\midrule
\multirow{5}{*}{Arxiv}
& Original     & 13.7 & 13{,}161 & 0.63 & 0.0  & 100.0 & 0.23 & 0.65 \\
& Random       & 9.6  & 9{,}263  & 0.64 & 5.0  & 92.7  & 0.14 & 0.65 \\
& K-Neighbor   & 6.4  & 4{,}702  & 0.44 & 0.0  & 100.0 & 0.11 & 0.64 \\
& Rank Degree  & 1.0  & 1{,}422  & 0.91 & 79.9 & 16.7  & 0.04 & 0.67 \\
& Local Degree & 6.3  & 13{,}161 & 0.57 & 0.0  & 100.0 & 0.25 & 0.62 \\
\midrule
\multirow{5}{*}{Products}
& Original     & 50.5 & 17{,}481 & 0.62 & 2.0  & 97.4 & 0.41 & 0.81 \\
& Random       & 46.0 & 15{,}916 & 0.62 & 2.4  & 97.0 & 0.37 & 0.81 \\
& K-Neighbor   & 8.5  & 2{,}414  & 0.37 & 2.0  & 97.4 & 0.09 & 0.83 \\
& Rank Degree  & 0.3  & 1{,}871  & 0.90 & 81.4 & 0.1  & 0.00 & 0.71 \\
& Local Degree & 12.2 & 17{,}481 & 0.64 & 2.0  & 97.4 & 0.36 & 0.78 \\
\midrule
\multirow{5}{*}{Papers100M}
& Original     & 29.1 & 251{,}471 & 0.67 & 0.0  & -- & -- & 0.57 \\
& Random       & 20.4 & 176{,}143 & 0.68 & 4.9  & -- & -- & 0.57 \\
& K-Neighbor   & 7.2  & 38{,}035  & 0.42 & 0.0  & -- & -- & 0.60 \\
& Rank Degree  & 0.5  & 62{,}354  & 0.87 & 75.7 & -- & -- & 0.50 \\
& Local Degree & 9.0  & 251{,}471 & 0.64 & 0.0  & -- & -- & 0.54 \\
\bottomrule
\end{tabular}
\end{table}

\end{document}